\documentclass[acmsmall]{acmart}
\AtBeginDocument{%
  }

\setcopyright{acmlicensed}
\copyrightyear{2026}
\acmYear{2026}
\acmDOI{XXXXXXX.XXXXXXX}

\acmJournal{TOMM}

\usepackage{graphicx}
\usepackage{array}
\usepackage{multirow}
\usepackage{multicol}
\usepackage{subcaption}

\begin{document}

\title{Towards Sustainable Universal Deepfake Detection with Frequency-Domain Masking}


\author{Chandler Timm C. Doloriel}
\authornote{The initial phase of this research was carried out while the first author was working for Singapore University of Technology and Design (SUTD).}
\orcid{0000-0003-3820-7317}
\affiliation{%
  \institution{Singapore University of Technology and Design (SUTD)}
  \city{Singapore}
  \country{Singapore}}
\affiliation{%
  \institution{Faculty of Science and Technology (REALTEK), Norwegian University of Life Sciences (NMBU)}
  \city{Aas}
  \country{Norway}}
\email{chandler.timm.cagmat.doloriel@nmbu.no}

\author{Habib Ullah}
\orcid{0000-0002-2434-0849}
\affiliation{%
  \institution{Faculty of Science and Technology (REALTEK), Norwegian University of Life Sciences (NMBU)}
  \city{Aas}
  \country{Norway}}
\email{habib.ullah@nmbu.no}

\author{Kristian Hovde Liland}
\orcid{0000-0001-6468-9423}
\affiliation{%
  \institution{Faculty of Science and Technology (REALTEK), Norwegian University of Life Sciences (NMBU)}
  \city{Aas}
  \country{Norway}}
\email{kristian.liland@nmbu.no}

\author{Fadi Al Machot}
\orcid{0000-0002-1239-9261}
\affiliation{%
  \institution{Faculty of Science and Technology (REALTEK), Norwegian University of Life Sciences (NMBU)}
  \city{Aas}
  \country{Norway}}
\email{fadi.al.machot@nmbu.no}

\author{Ngai-Man Cheung}
\authornote{corresponding author}
\orcid{0000-0003-0135-3791}
\affiliation{%
  \institution{Singapore University of Technology and Design (SUTD)}
  \city{Singapore}
  \country{Singapore}}
\email{ngaiman_cheung@sutd.edu.sg}

\renewcommand{\shortauthors}{Doloriel et al.}

\begin{abstract}
Universal deepfake detection aims to identify AI-generated images across a broad range of generative models, including unseen ones.
This requires robust generalization to new and unseen deepfakes—which emerge frequently—while minimizing computational overhead to enable large-scale deepfake screening, a critical objective in the era of Green AI.
%This requires robust generalization while minimizing computational overhead—a critical goal in the era of Green AI.
In this work, we explore frequency-domain masking as a training strategy for deepfake detectors. Unlike traditional methods that rely heavily on spatial features or large-scale pretrained models, our approach introduces random masking and geometric transformations, with a focus on frequency masking due to its superior generalization properties. We demonstrate that frequency masking not only enhances detection accuracy across diverse generators but also maintains performance under significant model pruning, offering a scalable and resource-conscious solution. Our method achieves state-of-the-art generalization on GAN- and diffusion-generated image datasets and exhibits consistent robustness under structured pruning. These results highlight the potential of frequency-based masking as a practical step toward sustainable and generalizable deepfake detection. Code and models are available at https://github.com/chandlerbing65nm/FakeImageDetection.
\end{abstract}




\received{18 May 2025}
\received[revised]{26 September 2025}
\received[accepted]{23 January 2026}

\maketitle

\section{Introduction}
  
\begin{figure}[t!]
    \centering
    \includegraphics[width=\linewidth]{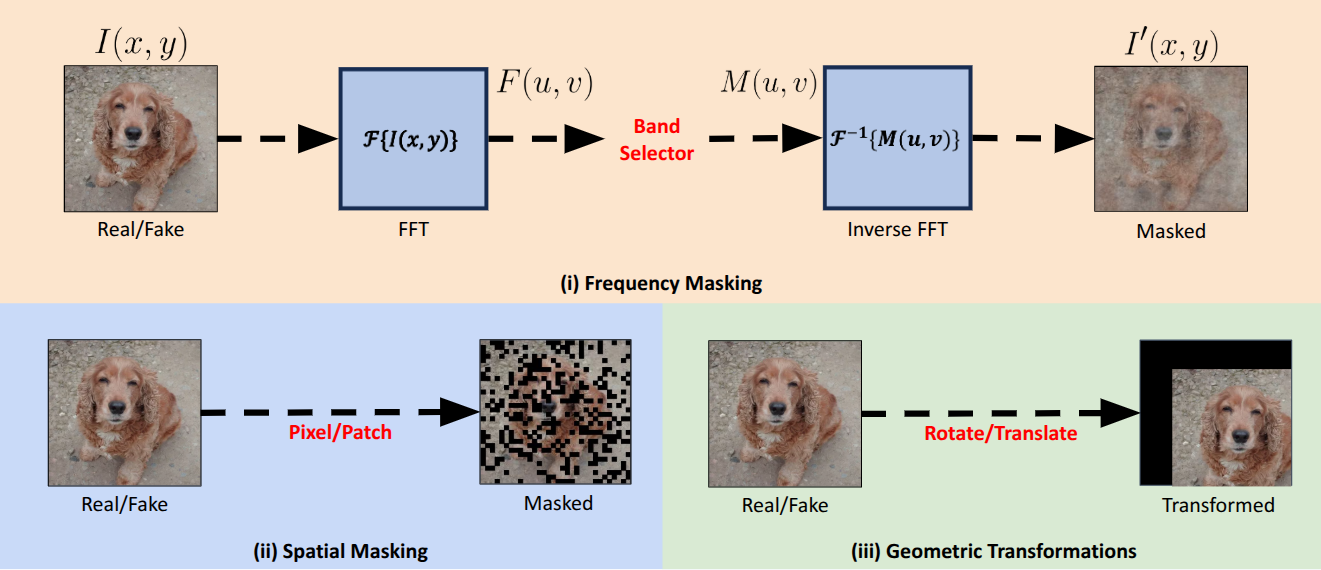}
    \caption{
        Our proposed training augmentation for universal deepfake detection using frequency masking is compared with spatial masking and geometric transformations.
        \textbf{(i) Frequency-domain masking}: The input image \( I(x, y) \) is transformed to the frequency domain \( F(u, v) \) via FFT. Guided by a frequency band selector, specific frequencies in \( F(u, v) \) are nullified to produce \( M(u, v) \). The inverse FFT yields the masked image \( I'(x, y) \), which trains
        the detector to enhance generation.
        %the classifier to suppress generator-specific spectral artifacts.
        \textbf{(ii) Spatial-domain masking}: The input image is masked at the pixel or patch level, occluding local regions while leaving frequency artifacts intact.
        \textbf{(iii) Geometric transformations}: The input undergoes spatial perturbations (e.g., translation), altering composition without affecting frequency-domain patterns.
        \textit{Remark}: Masking and transformations are applied \textit{only} during supervised training to encourage generalizable representations. They are not used during testing.
    }
    \label{fig_mainfig}
\end{figure}

The proliferation of increasingly convincing synthetic images, facilitated by generative AI, poses significant challenges across multiple sectors, including cybersecurity, digital forensics, and public discourse~\cite{ojha2023fakedetect, Gragnaniello2021, DBLP:conf/eccv/ChaiBLI20, Abdollahzadeh2023ASO}. These AI-generated images could be misused as deepfakes for malicious purposes, such as disinformation. Because these fake images can look very real, they can trick people and systems easily. Detection of deepfakes is an important problem that has attracted significant attention.

{\bf Universal deepfake detection.} Early research in deepfake detection primarily focused on identifying synthetic images generated by specific types of generative models, such as Generative Adversarial Networks (GANs)~\cite{mirsky2020deepfake}. However, with the rapid advancement of generative methods, including the rise of diffusion models~\cite{rombach2021diffusion}, the need for universal deepfake detection has become more prominent. This approach aims to detect deepfakes across a wide variety of generative AI models, including those that were not part of the training data for the detection system. Such a shift highlights the necessity for deepfake detection systems to exhibit robust generalization capabilities. In the early stages, Wang et al.~\cite{wang2019cnngenerated} explored the use of post-processing and data augmentation techniques to detect synthetic images, focusing particularly on those generated by GANs. Later, Chandrasegaran et al.~\cite{Chandrasegaran_2022_ECCV} introduced the concept of transferable forensic features, which are generalizable across different generative models, thus improving the detector’s ability to handle unseen models. More recent approaches, such as those by Ojha et al.~\cite{ojha2023fakedetect}, leverage the feature space of large pretrained models, showing strong performance across unseen generators, although they depend heavily on access to large-scale pretrained models. A notable advancement by Tan et al.~\cite{Tan2024RethinkingTU} investigated the impact of architectural artifacts, specifically those introduced by upsampling operations in generative models, and proposed detecting structural inconsistencies to improve generalization. Lastly, Chen et al.~\cite{Chen2022OSTIG} examined one-shot test-time training to enhance model generalization; however, this technique introduces a trade-off between accuracy and computational efficiency due to the need for additional resources per test sample.

{\bf The case for Green AI.} Universal deepfake detection, which aims for strong generalization across a variety of scenarios, often depends on large pre-trained models \cite{ojha2023fakedetect} and specialized  feature extraction techniques \cite{Tan2024FrequencyAwareDD}, both of which demand substantial computational resources. This dependency raises concerns about the scalability of these methods and their negative environmental impact due to the high energy consumption associated with training and inference. In response, the machine learning community is increasingly focusing on the concept of Green AI, which promotes the development of methods capable of achieving competitive performance while minimizing energy usage and computational cost \cite{chen2024gift,Kuo2022GreenLI,Zhu2022RGGIDAR,Zhu2021APixelHopAG,Wei2024AGL}. This shift in focus has led to efforts aimed at creating lightweight yet generalizable detection strategies that remain effective even in environments with limited resources. In this regard, frequency-domain representations stand out as an especially promising solution \cite{Coccomini2024DeepfakeDW}. These representations preserve important structural and statistical information, typically with fewer dimensions than their spatial counterparts, and require significantly lower computational overhead. Consequently, employing a frequency-based approach offers the potential to enhance both the generalization ability and efficiency of deepfake detection models. Beyond general-purpose detection, specialized domains (e.g., aquaculture) increasingly rely on synthetic data for training and augmentation. However, poor-quality synthetic samples may propagate errors in critical applications—from misdiagnosing fish diseases to inaccurate medical assessments. A robust universal detector must therefore generalize not only across generative architectures but also across diverse application domains with potentially limited training data.

{\bf Masked image modeling.} In recent years, masked image modeling (MIM) has gained significant attention as an effective self-supervised pre-training technique to enhance the generalization ability of models~\cite{DBLP:conf/cvpr/HeCXLDG22, DBLP:journals/corr/abs-2302-02615, DBLP:conf/iclr/0002LZ0OL23, DBLP:conf/nips/HuangCGXZLLX22}. This method revolves around pre-training models to predict missing or masked parts of unlabeled data, with reconstruction loss serving as the objective for learning. After the pre-training phase, the trained encoder can be effectively fine-tuned for a variety of downstream tasks, demonstrating strong transferability. He et al.~\cite{DBLP:conf/cvpr/HeCXLDG22} demonstrated that training with a masked autoencoder (MAE) allows high-capacity models to achieve state-of-the-art performance in terms of generalization across diverse applications. More recent works, such as those by Li et al.~\cite{DBLP:journals/corr/abs-2302-02615}, argue that reconstruction-based objectives excel at learning in-distribution representations, making them particularly effective for out-of-distribution detection. Xie et al.~\cite{DBLP:conf/iclr/0002LZ0OL23} expanded this approach by incorporating frequency-domain information into the masking process, proposing Masked Frequency Modeling (MFM), where the model learns to predict masked frequency components rather than spatial patches. This enables models to capture structured global image priors more efficiently. In the context of deepfake detection, Das et al.~\cite{Das2023UnmaskingDM} leveraged spatiotemporal transformers and masked autoencoding, showing that incorporating both spatial and temporal inconsistencies improves generalization across datasets.

{\bf In our work,} we propose to  explore masked image modeling to enhance the generalization capability of deepfake detectors with the objective of advancing universal deepfake detection. Unlike traditional masked image modeling, which primarily uses reconstruction loss in self-supervised pre-training, our method applies masking (Figure~\ref{fig_mainfig}) in a supervised setting, focusing on classification loss for distinguishing real and fake images. Our training involves both spatial and frequency domain masking on all images. This technique, which obscures parts of the image, enhances the challenge of training. It aims to prevent the detector from depending on superficial features and instead fosters the development of robust, generalizable representations. Importantly, masking is only employed during training, not in the testing phase.

We analyzed both random masking and geometric transformations for universal deepfake detection. Our results suggest that frequency masking is more effective in generalizing detection across diverse generative AI approaches. Our finding is consistent with a recent study by Corvi et al.~\cite{Corvi_2023_CVPR}, which identifies frequency artifacts in GAN- and diffusion-based synthetic images. Different from~\cite{Corvi_2023_CVPR}, our main contribution is a new training method that improves detection accuracy and robustness via frequency-domain masking. We further demonstrate that this method is resilient to model compression, i.e., structured pruning, making it attractive for real-world, resource-constrained deployments.

We remark that most existing detectors focus on spatial-domain artifacts~\cite{ojha2023fakedetect, Gragnaniello2021, wang2019cnngenerated}. Our contributions are summarized as follows:

\begin{enumerate}
    \item We present a study to explore masked image modeling in a supervised framework for universal deepfake detection.
    \item We analyze and compare two distinct types of masking methods (spatial and frequency) and two geometric transformation methods (rotation and translation), then empirically demonstrate that frequency masking offers better generalization (Figure~\ref{fig_mainfig}).
    \item We show that frequency masking as training augmentation preserves performance under model pruning, aligning with the principles of Green AI.
    \item We demonstrate that frequency masking improves universal detection performance across (a) standard GAN/diffusion benchmarks, and (b) specialized domains like aquaculture where synthetic data quality impacts critical decisions (e.g., fish health assessment).
\end{enumerate}
\section{Related Work}

In this section, we review prior work on universal deepfake detection, frequency-domain analysis, and masked image modeling, and discuss the approaches related to efficiency and Green AI, providing the context for our proposed frequency-domain masking strategy.

\subsection{Universal Deepfake Detection}

Early research in universal deepfake detection focused on identifying common artifacts in synthetic images. Zhang et al.~\cite{Zhang2019DetectingAS} proposed simulating GAN artifacts to train frequency-domain classifiers capable of generalizing across unseen models. Wang et al.~\cite{wang2019cnngenerated} discovered that CNN-generated images exhibit systematic artifacts detectable by CNN-based classifiers, even when the detector is trained on a single generator. This finding suggested shared weaknesses among early GAN models. In contrast, Gragnaniello et al.~\cite{Gragnaniello2021} emphasized the growing realism of synthetic content and highlighted the necessity of robust and generalizable detection tools in the face of increasingly sophisticated generative models. Chandrasegaran et al.~\cite{Chandrasegaran_2022_ECCV} introduced the idea of transferable forensic features, demonstrating that certain features—particularly color-based ones—can be learned to generalize across models. Ojha et al.~\cite{ojha2023fakedetect} proposed leveraging the feature space of large pretrained models for real-vs-fake classification. This method, without explicit model-specific training, was shown to outperform traditional classifiers on unseen generative models. More recently, Tan et al.~\cite{Tan2024RethinkingTU} explored architectural artifacts, particularly those introduced by upsampling operations in generative models. They proposed detecting structural inconsistencies such as neighboring pixel relationships (NPR), which yielded high generalization performance on unseen generators. Compared to these approaches, our method introduces masked image modeling as a training-time augmentation that reduces overfitting to generator-specific traces.

\subsection{Frequency-Domain Analysis}

Pioneering studies have explored frequency-domain cues as critical features for improving the robustness and generalizability of deepfake detection systems. Frank et al.~\cite{Frank2020LeveragingFA} first demonstrated that GAN-generated images often exhibit unnatural frequency artifacts due to upsampling operations, which can be effectively exploited to identify fakes. Qian et al.~\cite{Qian2020ThinkingIF} further investigated these frequency-aware clues, showing that subtle forgery traces and compression-induced anomalies manifest more clearly in the frequency domain, especially in low-quality images. Luo et al.~\cite{Luo2021GeneralizingFF} addressed the generalization challenge across forgery methods by proposing three modules to extract high-frequency noise patterns, which generalize better than spatial-domain textures. Complementary to this, Le and Woo~\cite{Le2021ADDFA} introduced a knowledge distillation framework where a student network learns from high-quality teacher signals in the frequency domain, enabling robust detection of compressed and degraded images. Jeong et al.~\cite{Jeong2022FrePGANRD} recognized the risk of overfitting to frequency-specific artifacts and proposed FrePGAN, which transitions from frequency-focused learning to incorporating image-level irregularities, thus balancing generalization across known and unknown generators. Tian et al.~\cite{Tian2023FrequencyAwareAF} combined RGB and frequency representations using attention-based fusion, showing that this dual-modality approach alleviates overfitting and enhances generalization. Most recently, Tan et al.~\cite{Tan2024FrequencyAwareDD} developed a frequency-aware architecture that targets high-frequency patterns across spatial and channel dimensions to improve generalizability while maintaining low model complexity. Li et al.~\cite{Li2024FreqBlenderED} introduced a data generation strategy that creates pseudo-fake images by blending frequency-domain features, enriching training data to better capture generalizable forgery traits. Distinctly, our approach employs frequency-domain masking as a supervised augmentation method that promotes learning of features resilient to domain shift.

\subsection{Masked Image Modeling}

Masked image modeling (MIM) has emerged as a powerful self-supervised learning paradigm in computer vision, with increasing relevance to generalization and robustness. He et al.~\cite{DBLP:conf/cvpr/HeCXLDG22} proposed Masked Autoencoders (MAE), demonstrating that reconstructing a high proportion of masked image patches using an asymmetric encoder-decoder architecture can achieve strong visual representations, rivaling or surpassing supervised pre-training on downstream tasks. Subsequently, Huang et al.~\cite{DBLP:conf/nips/HuangCGXZLLX22} introduced MaskedGAN, which uses randomized spatial and frequency masking to stabilize GAN training under data-scarce conditions, improving the robustness of generative models. Moving from representation learning to the domain of out-of-distribution detection, Li et al.~\cite{DBLP:journals/corr/abs-2302-02615} argued that reconstruction-based objectives are inherently better at learning in-distribution representations and thus more effective at detecting out-of-distribution samples, outperforming recognition-based methods. Xie et al.~\cite{DBLP:conf/iclr/0002LZ0OL23} extended MIM to the frequency domain with Masked Frequency Modeling (MFM), where the model learns to predict masked frequency components instead of spatial patches, capturing structured global image priors more efficiently. In the context of deepfake detection, Das et al.~\cite{Das2023UnmaskingDM} leveraged masked autoencoding with spatiotemporal transformers for videos, showing that masked modeling can facilitate strong generalization across datasets by encoding both spatial and temporal inconsistencies in fake content. More recently, Chen et al.~\cite{Chen2024MaskedCD} applied masking in conjunction with a conditional diffusion model to augment training data, demonstrating improved generalization to unseen deepfake forgeries. Building on these insights, our method incorporates random masking directly into supervised training, using frequency-domain masking to enforce feature robustness while maintaining low computational cost.

\subsection{Efficient and Green AI}

The increasing scale and complexity of deep neural networks have led to rising concerns over energy consumption and deployment cost, sparking interest in efficient and Green AI methods. Wen et al.~\cite{Wen2016LearningSS} addressed this early by introducing Structured Sparsity Learning (SSL), a method that enforces group-wise sparsity in network parameters to enable hardware-friendly acceleration without degrading performance. Liu et al.~\cite{Liu2017LearningEC} proposed Network Slimming, an approach that prunes unimportant channels via sparsity-induced scaling factors during training, achieving compact models compatible with standard hardware. Wang et al.~\cite{Wang2020NeuralPV} further advanced pruning techniques by introducing growing regularization to facilitate neural network compression without requiring expensive Hessian-based computations. Lee et al.~\cite{Lee2020LayeradaptiveSF} built on this with a layer-adaptive sparsity mechanism that assigns optimal sparsity levels across layers using a magnitude-based scoring strategy, improving the sparsity-performance tradeoff without manual tuning. Chen et al.~\cite{Chen2022DefakeHopAE} developed DefakeHop++, a lightweight deepfake detector that improves detection accuracy through expanded facial region coverage and supervised feature selection, outperforming CNN-based counterparts while maintaining a low parameter count. In a similar study, Vidya et al.~\cite{K2023CompressedDD} presented a spatio-temporal deepfake detector for compressed videos, employing model pruning and feature fusion to retain high performance with reduced model complexity. Fang et al.~\cite{Fang2023DepGraphTA} proposed DepGraph, a general structural pruning framework that models inter-layer dependencies to enable consistent and automatic pruning across various architectures, eliminating the need for manual design. Most recently, Lim et al.~\cite{Lim2024DistilDIREAS} proposed DistilDIRE, a diffusion-aware, lightweight deepfake detector that distills knowledge from large models to produce compact, fast models capable of accurate detection in real-time scenarios. Collectively, these works exemplify the trend towards efficient and scalable AI, aligning closely with our own method that emphasizes frequency-domain masking as a lightweight and pruning-resilient training strategy for sustainable universal deepfake detection~\cite{Kuo2022GreenLI,Zhu2021APixelHopAG,Zhu2022RGGIDAR,chen2024gift,Wei2024AGL}.
\section{Methodology}

\begin{figure}[t!]
\centering
\captionsetup[subfigure]{labelformat=empty}

% First row of images
\begin{subfigure}[b]{0.16\linewidth}
    \includegraphics[width=\linewidth]{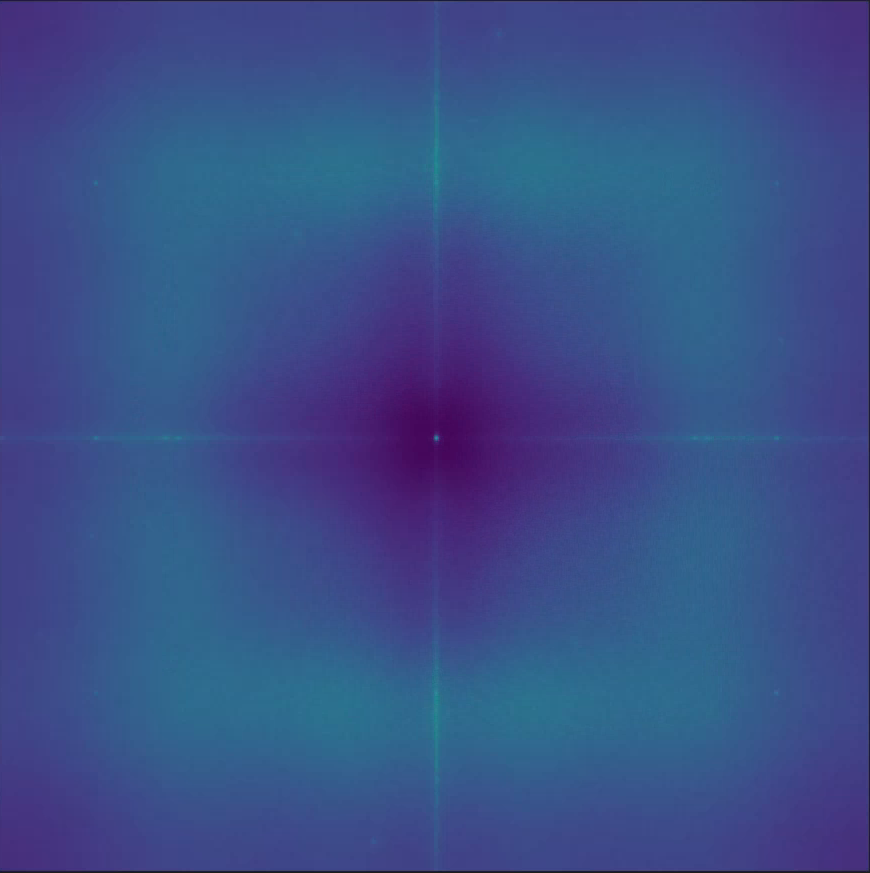}
    \label{fig:img1}
\end{subfigure}
\hfill
\begin{subfigure}[b]{0.16\linewidth}
    \includegraphics[width=\linewidth]{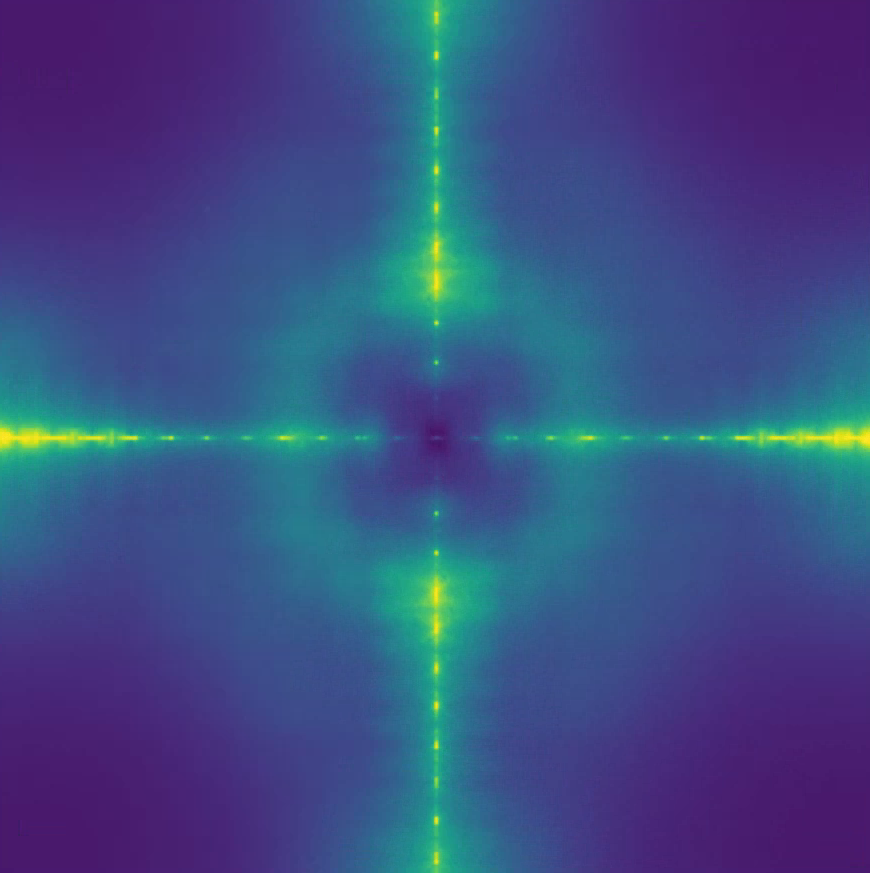}
    \label{fig:img2}
\end{subfigure}
\hfill
\begin{subfigure}[b]{0.16\linewidth}
    \includegraphics[width=\linewidth]{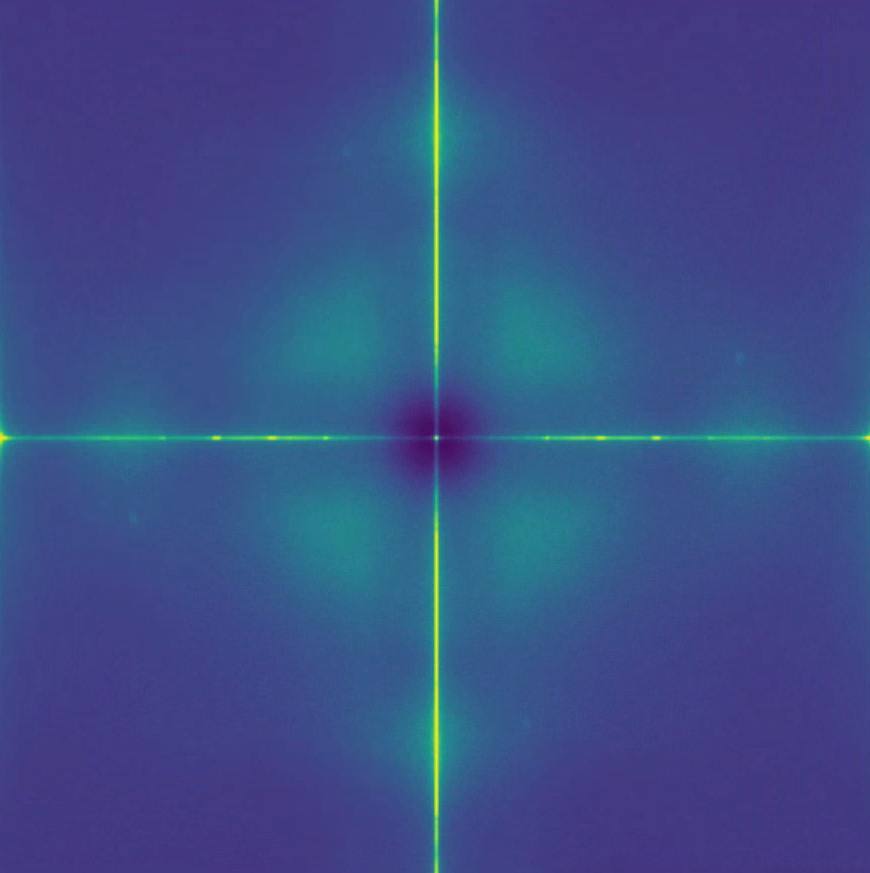}
    \label{fig:img3}
\end{subfigure}
\hfill
\begin{subfigure}[b]{0.16\linewidth}
    \includegraphics[width=\linewidth]{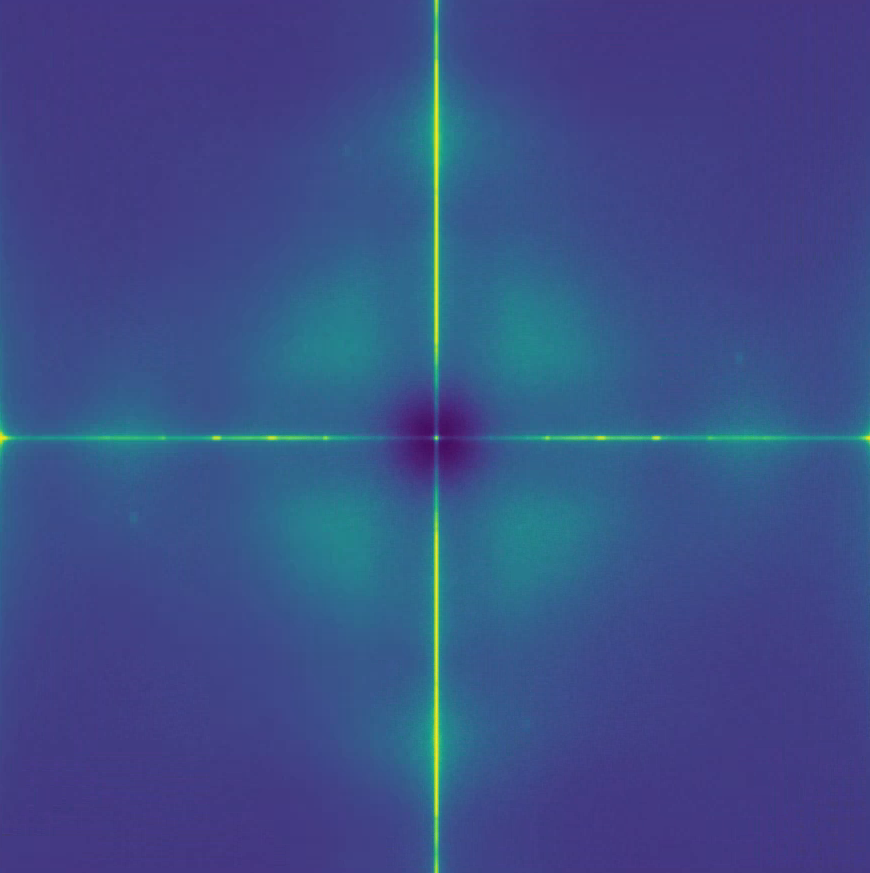}
    \label{fig:img4}
\end{subfigure}
\hfill
\begin{subfigure}[b]{0.16\linewidth}
    \includegraphics[width=\linewidth]{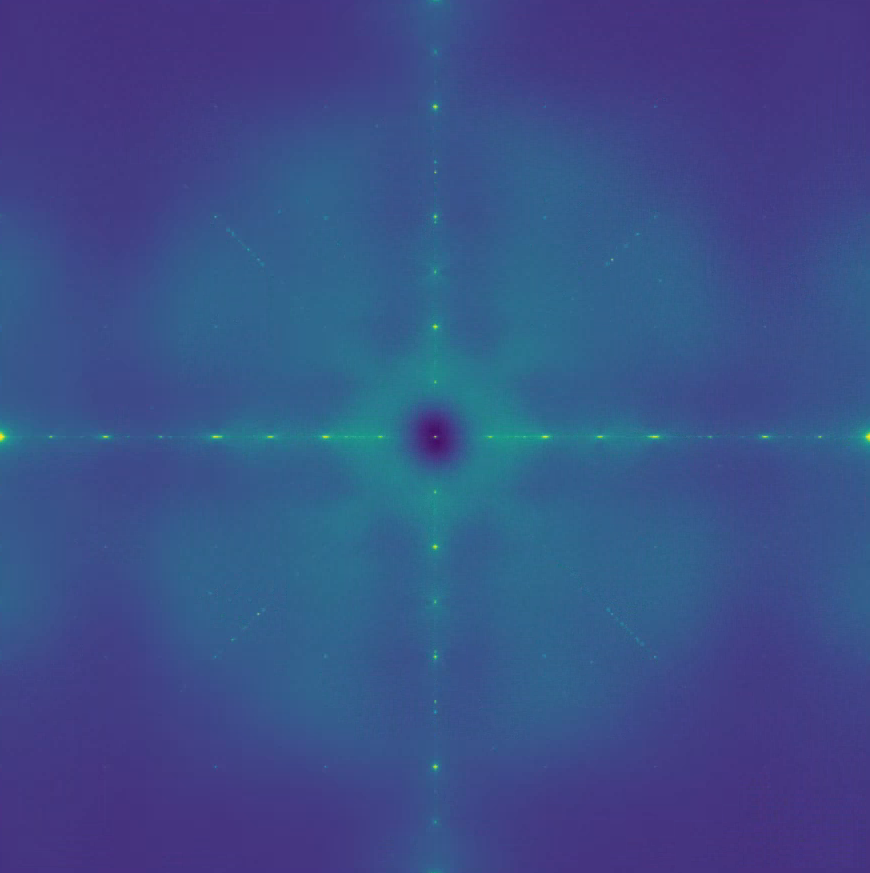}
    \label{fig:img5}
\end{subfigure}
\hfill
\begin{subfigure}[b]{0.16\linewidth}
    \includegraphics[width=\linewidth]{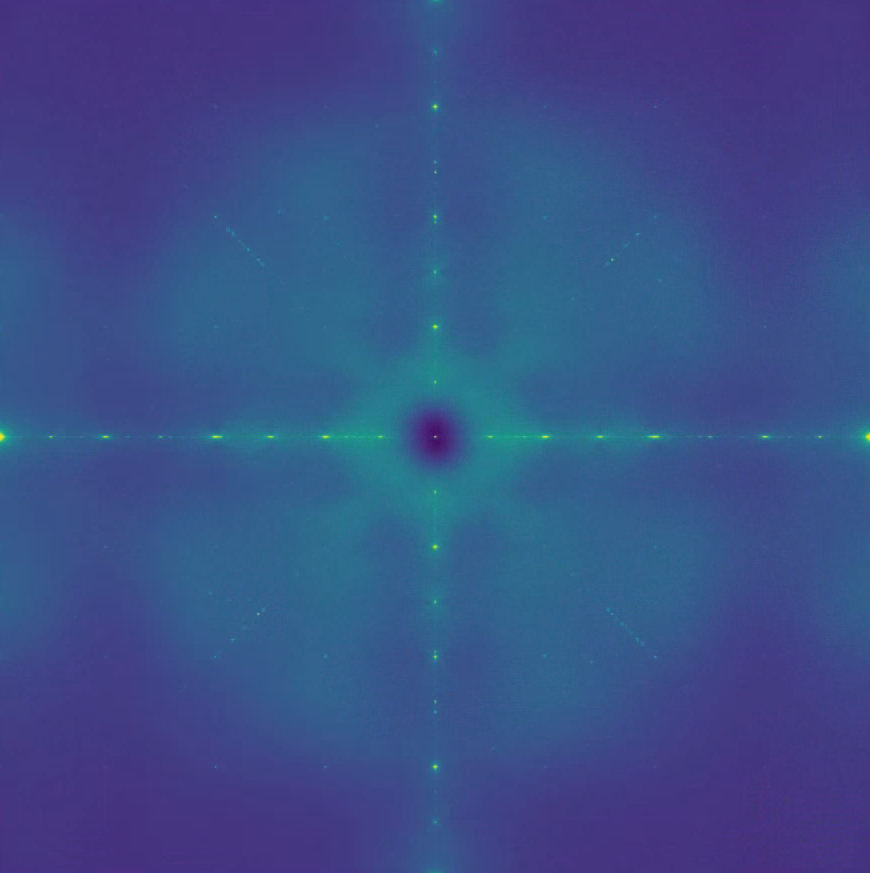}
    \label{fig:img6}
\end{subfigure}

% Second row of images
\begin{subfigure}[b]{0.16\linewidth}
    \includegraphics[width=\linewidth]{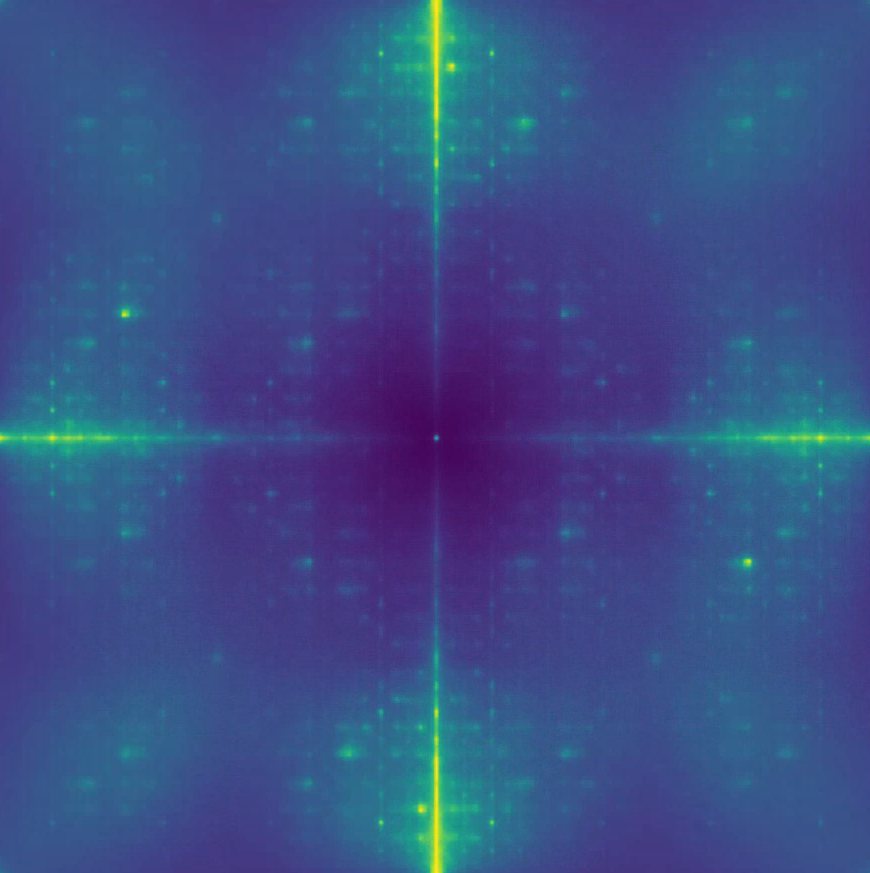}
    \caption{GauGAN}
    \label{fig:img7}
\end{subfigure}
\hfill
\begin{subfigure}[b]{0.16\linewidth}
    \includegraphics[width=\linewidth]{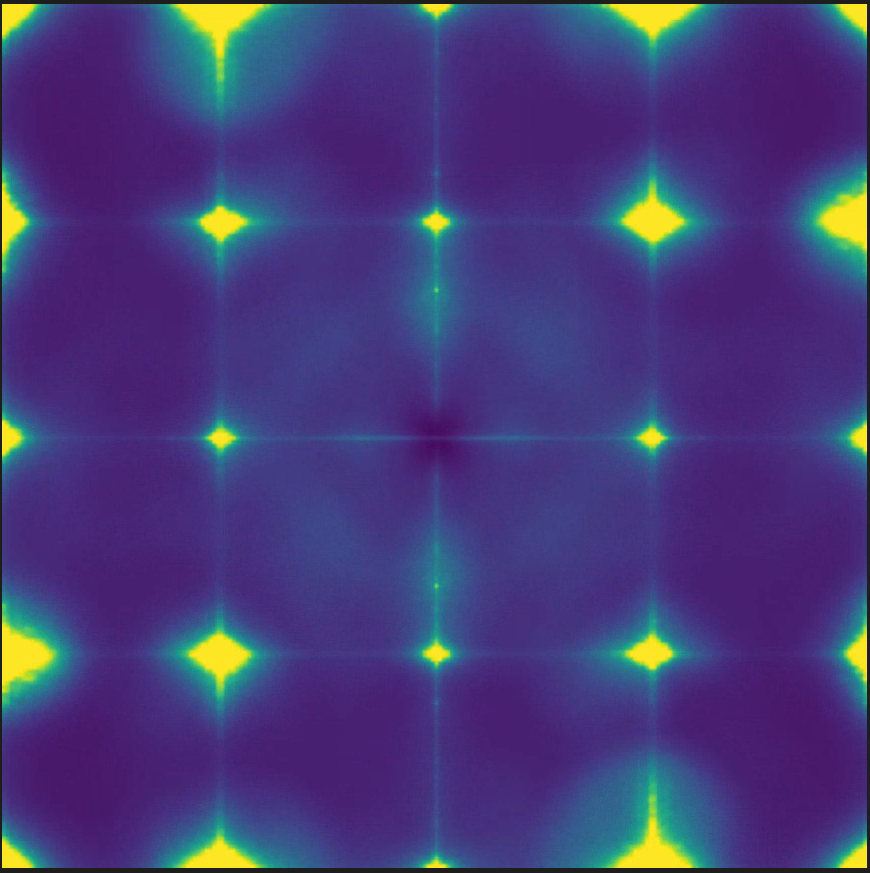}
    \caption{StarGAN}
    \label{fig:img8}
\end{subfigure}
\hfill
\begin{subfigure}[b]{0.16\linewidth}
    \includegraphics[width=\linewidth]{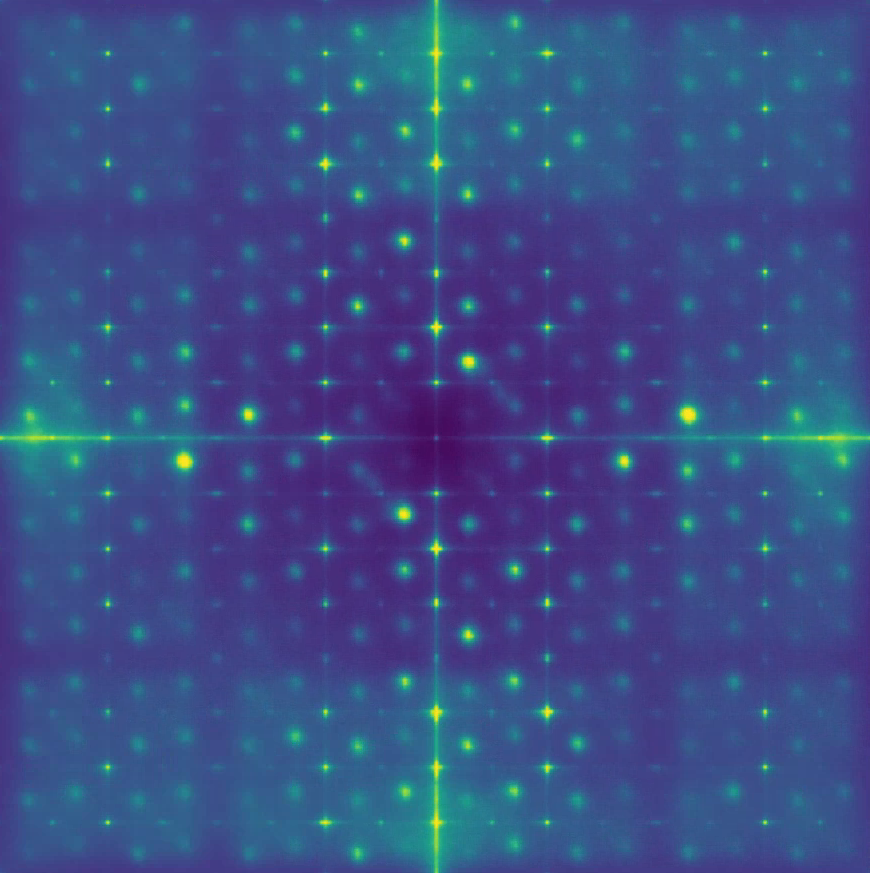}
    \caption{DALLE}
    \label{fig:img9}
\end{subfigure}
\hfill
\begin{subfigure}[b]{0.16\linewidth}
    \includegraphics[width=\linewidth]{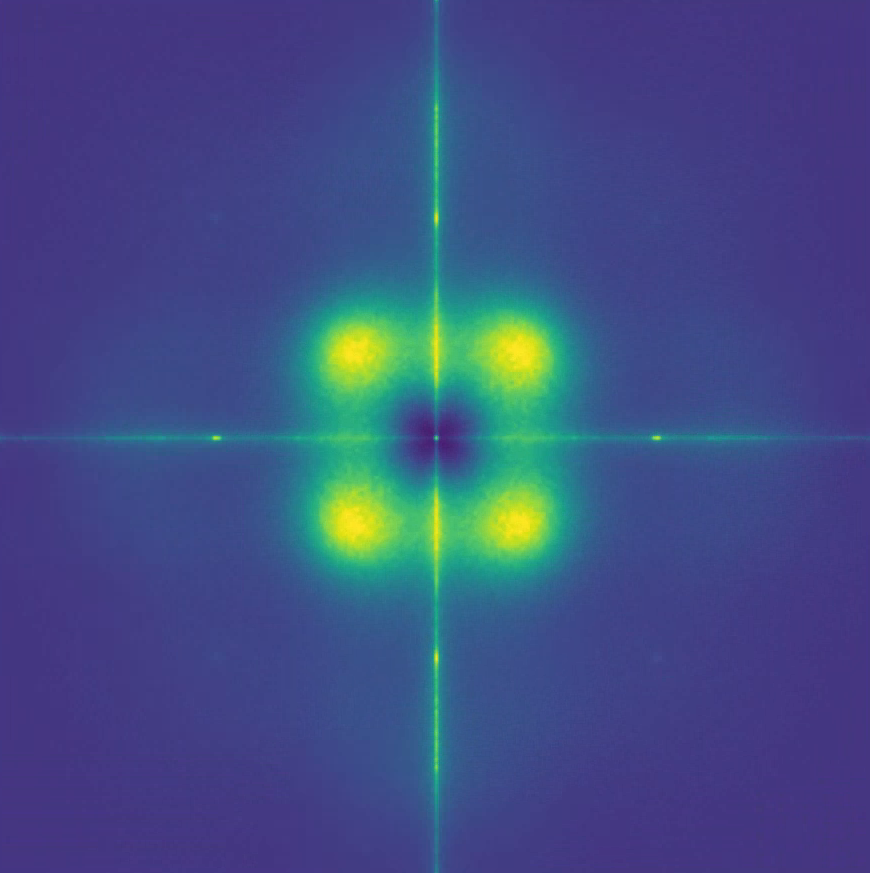}
    \caption{Glide}
    \label{fig:img10}
\end{subfigure}
\hfill
\begin{subfigure}[b]{0.16\linewidth}
    \includegraphics[width=\linewidth]{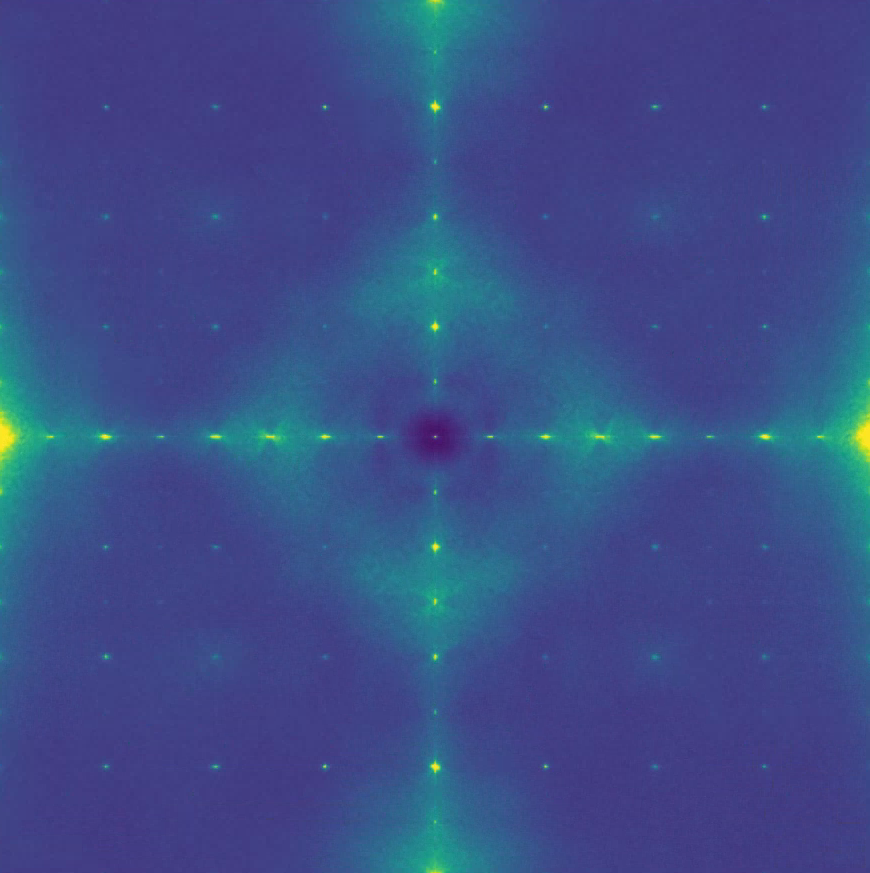}
    \caption{ControlNet}
    \label{fig:img11}
\end{subfigure}
\hfill
\begin{subfigure}[b]{0.16\linewidth}
    \includegraphics[width=\linewidth]{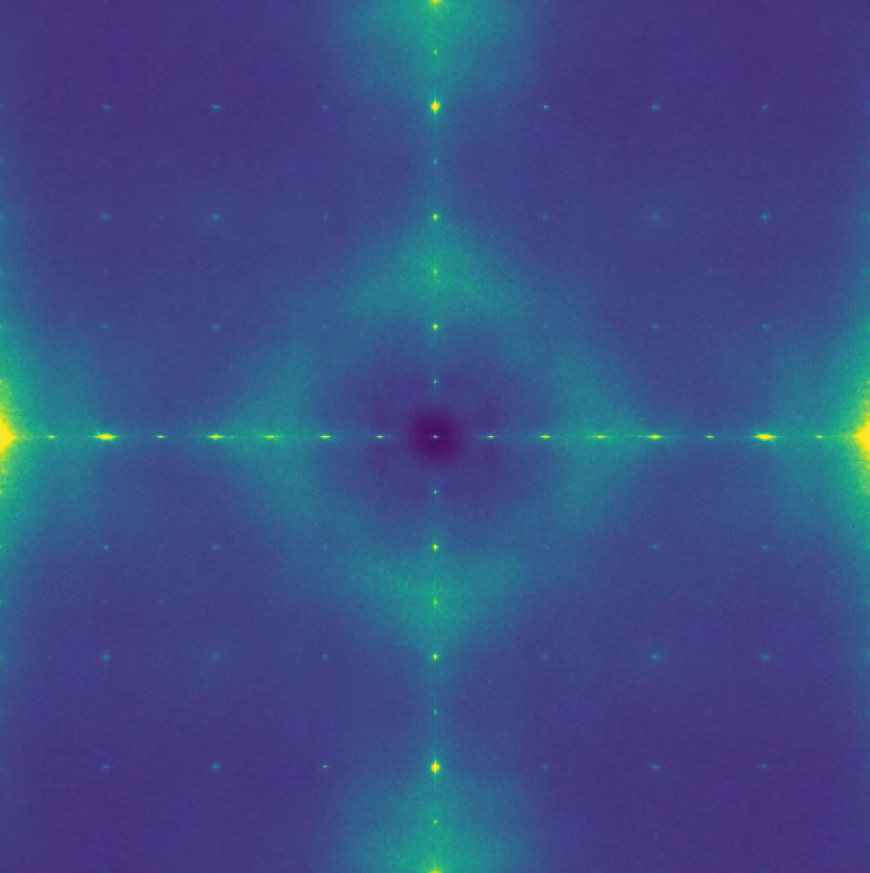}
    \caption{Stable Diffusion}
    \label{fig:img12}
\end{subfigure}
\caption{Comparison of frequency patterns from real images (top row) and AI-generated images (bottom row) using Fourier analysis. The images were processed to remove noise and standardized before analysis. Their frequency spectra were then averaged across multiple samples to highlight consistent patterns. The bottom row reveals distinct artificial patterns not found in real images, such as repetitive grid-like structures and unusual high-frequency patterns. The color intensity represents the strength of these frequency components, with brighter areas indicating stronger artificial signatures.
}
\label{fig_fftartifacts}
\end{figure}

To enhance the robustness and generalizability of universal deepfake detection systems, we propose and analyze masking strategies that operate in both spatial and frequency domains. Our methodology is centered around the hypothesis that masking portions of the input image during training compels the detection model to learn deeper and more transferable features. Notably, frequency analysis (Figure~\ref{fig_fftartifacts}) reveals that AI-generated images contain distinct artificial textures, such as repetitive grid-like structures and unusual high-frequency patterns, which are not found in real images. These characteristic periodic patterns differ across generative models and are clearly visible in the averaged frequency spectra, where brighter areas indicate stronger artificial signatures. This motivates targeted frequency-domain masking to suppress overfitting to such artifacts. 

These features, less reliant on surface-level artifacts, can be used to obtain better detection performance across a diverse set of synthetic image generators, including those unseen during training. This section details the specific design of the masking and geometric operations, implementation considerations, and their intended contributions to the model’s learning dynamics.

\subsection{Spatial Domain Masking}
\label{subsec:math_spatial_patch_masking}

Spatial domain masking operates directly on the pixel grid of the image and serves to obscure specific regions in the spatial layout. We employ two primary techniques under this category: \textbf{Patch Masking} and \textbf{Pixel Masking}. Each aims to encourage the model to learn contextual and global features rather than memorizing local or low-level visual patterns that may not generalize well to unseen generative models.

\textbf{Patch Masking} divides an input image of dimensions \(H \times W\) into a grid of non-overlapping patches, each of size \(p \times p\). The total number of patches, denoted as \(N\), is computed as \(N = \left\lfloor {H \times W}/{p^2} \right\rfloor\). A masking ratio \(r\) is specified to determine the proportion of patches to be masked. Accordingly, \(m = \lceil r \times N \rceil\) patches are selected at random, and all pixel values within these patches are set to zero, effectively occluding them from the model's view during training. This coarse masking strategy enforces the model to aggregate information from unmasked patches and infer useful representations despite partial occlusions.

\textbf{Pixel Masking}, on the other hand, applies a finer granularity of masking. Here, each individual pixel within the image is considered independently. Given a total of \(T = H \times W\) pixels in the image, we mask \(m = \lceil r \times T \rceil\) pixels, chosen randomly. These masked pixels are also set to zero, which results in a sparser but more stochastic occlusion pattern. Unlike patch masking, pixel masking introduces localized perturbations throughout the image, potentially disrupting low-level patterns and forcing the model to develop resilience against such noise.

The masking operation common to both strategies can be mathematically described as:

\begin{equation}
M(x, y) = \begin{cases}
0 & \text{if }(x, y) \in m \\
I(x, y) & \text{otherwise}
\end{cases}
\end{equation}

\noindent where \(M(x, y)\) represents the pixel value at location \((x, y)\) in the masked image, \(I(x, y)\) denotes the original image's pixel value at the same location, and \(m\) is the set of pixels (or patch centers) selected for masking. This binary masking operation is element-wise multiplied with the input image to produce the masked image used for training. Importantly, this masking is applied only during training.

These spatial masking strategies serve as a baseline for understanding the effect of domain-specific occlusions. They provide valuable insights into how different masking granularities impact the model's ability to generalize, especially when detecting forgeries crafted by unseen generative models.

\subsection{Frequency Domain Masking}

In contrast to spatial masking, frequency domain masking operates on the transformed representation of the image in the Fourier domain. The motivation for frequency-domain masking lies in the observation that many generative models, especially those involving upsampling operations, introduce artifacts that manifest distinctly in the frequency spectrum. By selectively masking frequency components, we aim to suppress reliance on such generator-specific artifacts and promote the learning of more global and intrinsic visual features.

We begin by transforming the input image from the spatial domain to the frequency domain using the Fast Fourier Transform (FFT). Let \(I(x, y)\) denote the original image, where \(x\) and \(y\) are spatial coordinates. The corresponding frequency representation \(F(u, v)\) is computed as:

\begin{equation}
F(u, v) = \mathcal{F}\{ I(x, y) \} = \sum_{x=0}^{H-1} \sum_{y=0}^{W-1} I(x, y) \cdot e^{-2\pi i \left( \frac{ux}{H} + \frac{vy}{W} \right)}
\end{equation}

Here, \(F(u, v) = \mathcal{F}\{ I(x, y) \}\) denotes the 2D FFT operation, where \(u\) and \(v\) correspond to the frequency indices along the vertical and horizontal axes, respectively. The output \(F(u, v)\) is a complex-valued matrix representing the amplitude and phase of each frequency component.

We define four types of masking based on specific frequency bands (based on heuristics):
\begin{itemize}
  \item \textbf{Low Band:} Corresponds to low-frequency components where most of the global image structure resides. Defined as \(0 \leq u < \frac{H}{4}, 0 \leq v < \frac{W}{4}\).
  \item \textbf{Mid Band:} Contains intermediate frequency components, capturing medium-level textures and transitions. Defined as \(\frac{H}{4} \leq u < \frac{3H}{4}, \frac{W}{4} \leq v < \frac{3W}{4}\).
  \item \textbf{High Band:} Comprises high-frequency components often associated with fine details and noise. Defined as \(\frac{3H}{4} \leq u < H, \frac{3W}{4} \leq v < W\).
  \item \textbf{All Band:} Represents the entire frequency spectrum, with masking applied randomly across all frequency components.
\end{itemize}

Given a masking ratio \(r\), we randomly select \(r \times B\) frequency bins within the designated band \(B\) (low, mid, high, or all) to be masked. These selected frequency components are zeroed out in the amplitude spectrum, while the phase information is retained to preserve the spatial consistency of the image. The masked frequency representation is then inverse-transformed back to the spatial domain using the Inverse FFT (IFFT), yielding a corrupted image that is used as input during training.

This process can be formally described as:

\begin{equation}
M(u, v) = \begin{cases}
0 & \text{if }(u, v) \in \mathcal{B}_r \\
F(u, v) & \text{otherwise}
\end{cases}
\end{equation}

\noindent where \(\mathcal{B}_r\) is the set of frequency indices in band \(B\) selected for masking according to ratio \(r\). The inverse FFT of \(M(u, v)\) reconstructs the masked image \(I'(x, y)\), which exhibits partial frequency loss.

Frequency-domain masking encourages the detector to rely less on artifact-heavy signals concentrated in particular frequency bands, especially those associated with synthetic image generation pipelines. By suppressing specific frequency bands during training, we facilitate the development of frequency-invariant representations, which are crucial for generalization across various generative models.

\subsection{Geometric Transformations}
\label{subsec:geometric_augmentations}

In addition to spatial and frequency domain masking, we introduce two additional geometric augmentations—\textbf{rotation} and \textbf{translation}—to further compare the model’s ability to generalize across different augmentations. These augmentations are applied to the input image with the goal of increasing the model's robustness to variations in object pose and spatial shifts, which are common in real-world applications.

\textbf{Rotation Augmentation} involves applying a random rotation angle \(\theta\) within a predefined range. Specifically, the rotation angle is sampled uniformly from the interval \([0, r\times180^\circ]\), and the input image \(I(x, y)\) is rotated by the chosen angle \(\theta\), where \(r\) is the ratio of rotation angle to choose. The rotation operation can be expressed as:

\begin{equation}
I'(x, y) = R_{\theta}(I(x, y)),
\end{equation}

\noindent where \(R_{\theta}(I(x, y))\) is the rotated image, and \(\theta\) is the random angle selected. The rotation matrix \(R_{\theta}\) transforms the pixel coordinates as follows:

\begin{equation}
\begin{pmatrix}
x' \\
y'
\end{pmatrix}
= 
\begin{pmatrix}
\cos(\theta) & -\sin(\theta) \\
\sin(\theta) & \cos(\theta)
\end{pmatrix}
\begin{pmatrix}
x \\
y
\end{pmatrix},
\end{equation}

\noindent where \((x, y)\) are the original pixel coordinates, and \((x', y')\) are the new coordinates after rotation.

\textbf{Translation Augmentation} applies random translations to the image, shifting it horizontally and/or vertically by a random fraction of the image dimensions. Specifically, we define a translation ratio \(r\), where the horizontal translation is a random shift between \(dx = [-r \times W, r \times W]\) and the vertical translation is between \(dy = [-r \times H, r \times H]\), with \(W\) and \(H\) representing the width and height of the image. The translation operation is then:

\begin{equation}
I'(x, y) = T_{(dx, dy)}(I(x, y)),
\end{equation}

\noindent where \(T_{(dx, dy)}\) represents the translation operation with displacement \(dx\) and \(dy\), which are random values sampled within the specified translation range. The translation operation can be modeled as:

\begin{equation}
(x', y') = (x + dx, y + dy),
\end{equation}

Both rotation and translation augmentations are applied during training, which encourages the model to learn pose and shift-invariant features, thus improving its generalization performance when detecting deepfakes from unseen generators.

\subsection{Structured Layer-wise Pruning}

To investigate the relationship between model capacity and augmentation effectiveness, we employ structured pruning to systematically reduce network parameters. For a convolutional layer with weight tensor $W \in \mathbb{R}^{C_{out} \times C_{in} \times K \times K}$, where $C_{out}$ and $C_{in}$ represent output and input channels respectively, and $K$ is the kernel size, we define the pruning operation as:

\begin{equation}
W_{pruned} = W \odot M,\quad M_{i} = \begin{cases} 
1 & \text{$C_{out}$ at index } i \text{ is retained} \\
0 & \text{otherwise}
\end{cases}
\end{equation}

By default, structured pruning removes \textbf{output channels} in each convolutional layer based on their importance. The importance scores $s \in \mathbb{R}^{C_{out}}$ are computed using the L1-norm of the corresponding output channel weights:

\begin{equation}
s_i = \|W_i\|_1 = \sum_{j=1}^{C_{in}} \sum_{k_1=1}^{K} \sum_{k_2=1}^{K} |W_{i,j,k_1,k_2}|
\end{equation}

For a target pruning ratio $p \in \{0\%, 20\%, 50\%, 80\%\}$, we retain the top $(1-p)\times C_{out}$ output channels according to $s$. Additionally, to maintain architectural consistency, we \textbf{prune the corresponding input channels in the subsequent layer if dependencies require it}. That is, if an output channel is removed in one layer, the matching input channel in the next connected layer must also be pruned to preserve valid dimensionality.

The pruning procedure is formalized as:

\begin{equation}
\mathcal{P}(W,p) = \{W_i : s_i \geq \tau_p\},\quad \tau_p = \text{top-}(1-p)\text{ quantile of } s
\end{equation}

After training with augmentations, we apply $\mathcal{P}$ to each convolutional layer and fine-tune the pruned model for 5 epochs using 2\% of the ProGAN training data. This procedure enables analysis of augmentation benefits across varying model capacities while preserving structural compatibility through coordinated pruning of output channels (and input channels when required by inter-layer dependencies). The final evaluation uses the full test set without augmentation to measure the preserved detection capability under compression.

\section{Experiments}

\textbf{Datasets}: Our experiments adhere to the training/validation setup of Wang et al.~\cite{wang2019cnngenerated}, using ProGAN (720k training and 4k validation samples). For testing, we evaluate on the following datasets:

\begin{itemize}
    \item Wang et al.~\cite{wang2019cnngenerated}: 11 datasets covering GANs (ProGAN, StyleGAN, BigGAN, GauGAN, CycleGAN, StarGAN), perceptual-loss models (CRN, IMLE), low-level vision (SITD, SAN), and DeepFake images.
    \item Ojha et al.~\cite{ojha2023fakedetect}: 8 diffusion model datasets including Guided Diffusion, Latent Diffusion (LDM) with varying steps of noise refinement (e.g., 100, 200) and generation guidance (w/ CFG), Glide which used two stages of noise refinement steps. The first stage (e.g., 50, 100) is to get low-resolution 64 × 64 images and then use 27 steps to upsample the image to 256 × 256 in the next stage, and lastly DALL-E-mini.
    \item Das et al.~\cite{Das2024BDfreshwaterfishAI}: 2 synthetic fish datasets (ControlNet and Stable Diffusion) at 512x512 resolution with neutral backgrounds, avoiding underwater scenes.
\end{itemize}

\textbf{State-of-the-art (SOTA)}: 
We compared with following methods: 
Chai et al.~\cite{DBLP:conf/eccv/ChaiBLI20}, Natraj et al.~\cite{Nataraj2019DetectingGG}, Zhang et al.~\cite{Zhang2019DetectingAS}, Ojha et al.~\cite{ojha2023fakedetect}, Tan et al.~\cite{Tan2024RethinkingTU}, Wang et al.~\cite{wang2019cnngenerated} and Gragnaniello et al.~\cite{Gragnaniello2021}. Wang et al.~\cite{wang2019cnngenerated} and Gragnaniello et al.~\cite{Gragnaniello2021} methods achieved SOTA detection performance across GANs and diffusion models according to~\cite{Corvi_2023_ICASSP}. We applied our proposed frequency masking on these SOTA and evaluated the improvement. 

\textbf{Data processing}:
During training, we simulate various image post-processing operations by training all models with randomly left-right flipped and 224-pixel cropped images, while evaluating several augmentation variants: (1) Blur+JPEG (0.5): 50\% probability for Gaussian blur and JPEG, and (2) Blur+JPEG (0.1): same as (1) but with 10\% probability. These augmentation strategies help improve model robustness against common image manipulations encountered in real-world scenarios. The specific probabilities were chosen based on the work of Wang et al.~\cite{wang2019cnngenerated} showing their effectiveness in balancing regularization strength and preservation of original image features.

\textbf{Evaluation}:
Following other works~\cite{wang2019cnngenerated,ojha2023fakedetect,Gragnaniello2021}, we evaluate our model's performance on each dataset separately using mean average precision (mAP) and Area Under Receiver Operating Curve (AUROC), as these threshold-less, ranking-based scores are not sensitive to the threshold of real versus fake images and we expect performance to depend on each dataset's semantic content. During testing, each image is center-cropped to 224$\times$224 pixels without resizing to match the training post-processing pipeline, with no data augmentation applied. This consistent evaluation protocol ensures fair comparison across different models and datasets. Additionally, we report both metrics to provide comprehensive insights into model performance across different operational points.

As our default setting, we used Wang et al.~\cite{wang2019cnngenerated} and incorporated image augmentations like Gaussian Blur and JPEG compression, each having a 10\% likelihood of being applied. This configuration was selected as it provided the best trade-off between generalization performance and computational capacity in our experiments.

\subsection{Comparison of Augmentation Types}

\begin{figure}[t!]
\centering
\includegraphics[width=0.75\linewidth]{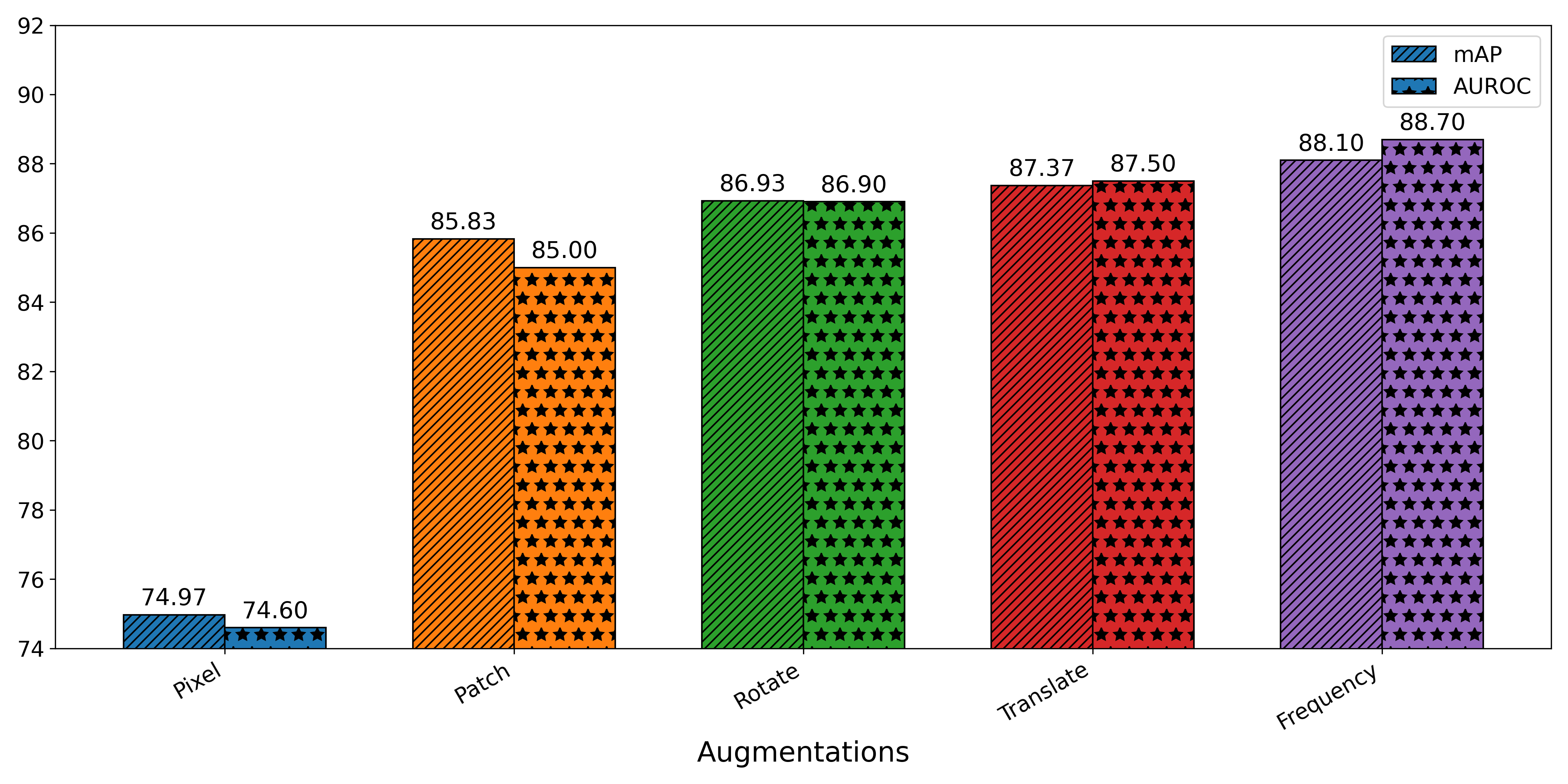}
\caption{The performance of different augmentation types is evaluated in terms of mean Average Precision (mAP) and Area Under the Receiver Operating Characteristic Curve (AUROC). Pixel, patch, and frequency masking are applied with a 15\% masking ratio, while geometric transformations use 27$^\circ$ (15\% of 180$^\circ$) random rotation and $\pm$15\% translation. Among these, frequency-based masking achieves the highest mAP and AUROC, demonstrating its superior effectiveness compared to other masking approaches.}
\label{fig_masking_types}
\end{figure}

The performance of different augmentation strategies, as illustrated in Figure~\ref{fig_masking_types}, reveals a clear hierarchy in detection performance. Pixel masking, the most basic approach, yields the lowest performance, suggesting that simple pixel-level perturbations may not sufficiently disrupt deepfake artifacts to improve generalization. Patch masking, which operates on larger 8x8 pixel blocks, demonstrates a significant improvement over pixel masking, reinforcing the idea that structured, localized modifications are more effective in exposing synthetic patterns. However, the highest performance is achieved by frequency-based masking, which surpasses both pixel and patch masking, indicating that manipulating frequency-domain features captures more discriminative and generalizable traces of deepfake manipulation. This aligns with prior work suggesting that frequency analysis is highly effective in detecting synthetic media, as many generative models leave subtle but detectable artifacts in the frequency spectrum. Additionally, geometric augmentations (rotation and translation) showed better performance than spatial masking but lags behind frequency masking. Overall, these results highlight the importance of selecting appropriate augmentations, with frequency-based emerging as the most effective for universal deepfake detection.

\subsection{Ratio of Frequency Masking}

\begin{table}[t!]
    \centering
    \resizebox{0.5\linewidth}{!}{\tiny
    \begin{tabular}{c|c|c}
        \hline
        Masking Ratio (\%) & mAP & AUROC \\
        \hline
        0\%  & 85.11            & 82.99 \\
        15\% & \textbf{88.10}   & \textbf{88.72} \\
        30\% & 87.24            & 87.58 \\
        50\% & 85.31            & 86.13 \\
        70\% & 83.86            & 84.10 \\
        \hline
    \end{tabular}}
    \vspace{1em}
    \caption{The table presents mAP and AUROC scores for varying degrees of frequency masking. The highest values for both metrics are achieved at a 15\% masking ratio.}
    \label{tab_maskratios}
\end{table}

As shown in Table \ref{tab_maskratios}, we observe a clear trend in the performance of our frequency masking technique across different masking ratios. The highest mean average precision (mAP) of \textbf{88.10\%} and AUROC of \textbf{88.72\%} are achieved at a masking ratio of \textbf{15\%}. Notably, the AUROC scores closely align with the mAP trends, peaking at 15\% masking and gradually declining as the masking ratio increases. This consistency across metrics reinforces that the model is sensitive to the proportion of frequencies being masked. Excessive masking (e.g., 70\%) not only reduces mAP by 4.24 percentage points but also lowers AUROC by 4.62 points compared to the 15\% baseline. suggesting that aggressive masking compromises the model's ability to distinguish between positive and negative samples. Therefore, based on these results, we select \textbf{15\%} as the default masking ratio for frequency masking in our experiments.

\subsection{Frequency Bands for Masking}

\begin{table}[t!]
\label{table:combined}
\centering
\begin{tabular}{ccccc|ccccc}
\hline
\multicolumn{1}{c}{\multirow{2}{*}{\parbox{1.5cm}{\centering Generative Models}}} & \multicolumn{4}{c}{mAP} & \multicolumn{4}{c}{AUROC} \\
\cline{2-9}
\multicolumn{1}{c}{} & Low & Mid & High & All & Low & Mid & High & All \\
\hline
 GANs             & 95.7 & 93.9 & 94.9 & \textbf{96.2} & 96.0 & 94.8 & \textbf{95.1} & 96.1 \\
 DeepFake         & 85.6 & \textbf{87.2} & 80.2 & 79.1 & \textbf{88.0} & 86.4 & 82.3 & 80.5 \\
 Low-Level Vision & 85.8 & 83.7 & \textbf{88.8} & 87.2 & 84.5 & 83.0 & \textbf{88.9} & 85.5 \\
 Perceptual Loss  & \textbf{99.2} & 98.0 & 98.1 & 98.4 & \textbf{99.2} & 98.2 & 98.3 & 98.6 \\
 Guided Diffusion & \textbf{74.8} & 70.2 & 69.4 & 72.6 & \textbf{74.8} & 71.8 & 70.1 & 74.0 \\
 Latent Diffusion & 75.3 & 76.0 & 66.9 & \textbf{80.1} & 77.3 & 77.1 & 70.1 & \textbf{81.8} \\
 Glide            & \textbf{84.8} & 81.1 & 74.1 & 84.7 & 86.1 & 81.5 & 76.6 & \textbf{86.3} \\
 DALL-E           & 69.4 & 69.2 & 71.6 & \textbf{79.7} & 70.8 & 67.4 & 72.8 & \textbf{82.0} \\
\hline
 average      & 87.1 & 85.5 & 83.6 & \textbf{88.1} & 87.8 & 85.9 & 84.8 & \textbf{88.7} \\
\hline
\end{tabular}
\vspace{1em}
\caption{mAP and AUROC across various datasets from generative models masked in different frequency bands.}
\label{tab_freqbands}
\end{table}

Table \ref{tab_freqbands} presents both mAP and AUROC scores across various generative models when different frequency bands---Low, Mid, High, and All---are randomly masked. The results demonstrate that masking \textbf{all frequency bands} consistently yields the best overall performance, achieving the highest average mAP (\textbf{88.1\%}) and AUROC (\textbf{88.7\%}) scores. This comprehensive masking strategy appears most effective for universal deepfake detection, as it likely captures artifacts across the entire frequency spectrum. However, the analysis reveals interesting model-specific behaviors: while GANs and DALL-E show optimal performance with All-band masking, other models like DeepFake and Low-Level Vision achieve their best results with Mid-frequency (\textbf{87.2\%} mAP) and High-frequency (\textbf{88.8\%} mAP, \textbf{88.9\%} AUROC) masking respectively. Similarly, Perceptual Loss models perform best with Low-frequency masking (\textbf{99.2\%} for both metrics). These variations suggest that different generative models leave distinct forensic artifacts in specific frequency bands, potentially enabling targeted detection approaches. Nevertheless, the superior average performance of All-band masking underscores its robustness as a general-purpose strategy for detecting diverse deepfake generation methods.

Furthermore, masking low frequencies tends to help generalization because it prevents the model from leaning on broad, dataset-specific cues, and instead encourages it to learn the mid–high frequency details (e.g., subtle upsampling or periodic artifacts) that are more indicative of synthesis. By contrast, masking high frequencies during training removes exactly those fine-grained forensic signals the detector needs to learn, which can hurt performance. This means that in deepfake detection they are often more discriminative, while low-frequency content mostly carries semantics and potential shortcuts.

\subsection{Comparison with State-Of-The-Art}

\begin{table}[t]
\renewcommand{\arraystretch}{1.7}  % Adjust the row height
\centering
\Huge
\resizebox{\linewidth}{!}{
\begin{tabular}{cc|ccccccccccccccccccc|c}
\hline
\multirow{2}{*}{Method} & \multirow{2}{*}{Backbone} & \multicolumn{6}{c}{Generative Adversarial Networks} & \multirow{2}{*}{\parbox{1.5cm}{\centering  Deep\\Fake}} & \multicolumn{2}{c}{Low-Level Vision} & \multicolumn{2}{|c}{Perceptual Loss} & \multirow{2}{*}{\parbox{2.2cm}{\centering Guided}} & \multicolumn{3}{c|}{LDM} & \multicolumn{3}{c}{Glide} & \multirow{2}{*}{\parbox{2.8cm}{\centering DALL-E}} & average \\
\cline{3-8} \cline{10-13} \cline{15-20}
  &  & \parbox{1.5cm}{\centering Pro\\GAN} & \parbox{1.5cm}{\centering Cycle\\GAN} & \parbox{1.5cm}{\centering Big\\GAN} & \parbox{1.5cm}{\centering Style\\GAN} & \parbox{1.5cm}{\centering Gau\\GAN} & \parbox{1.5cm}{\centering Star\\GAN} &  & \parbox{1.5cm}{\centering SITD} & \parbox{1.5cm}{\centering SAN} & \parbox{1.5cm}{\centering CRN} & {\centering IMLE} &  & \parbox{1.5cm}{\centering 200\\steps} & \parbox{2.4cm}{\centering 200\\w/ CFG} & \parbox{1.5cm}{\centering 100\\steps} & \parbox{1.5cm}{\centering 100\\27} & \parbox{1.5cm}{\centering 50\\27} & \parbox{1.5cm}{\centering 100\\10} &  & mAP \\
\hline
  \multirow{2}{*}{Chai et al.~\cite{DBLP:conf/eccv/ChaiBLI20}} & ResNet50 & 98.86 & 72.04 & 68.79 & 92.96 & 55.9 & 92.06 & 60.18 & 65.82 & 52.87 & 68.74 & 67.59 & 70.05 & 87.84 & 84.94 & 88.1 & 74.54 & 76.28 & 75.84 & 77.07 & 75.28 \\
  & Xception & 80.88 & 72.84 & 71.66 & 85.75 & 65.99 & 69.25 & 76.55 & 76.19 & 76.34 & 74.52 & 68.52 & 75.03 & 87.1 & 86.72 & 86.4 & 85.37 & 83.73 & 78.38 & 75.67 & 77.73 \\

  Natraj et al.~\cite{Nataraj2019DetectingGG} & - & 99.74 & 80.95 & 50.61 & 98.63 & 53.11 & 67.99 & 59.14 & 68.98 & 60.42 & 73.06 & 87.21 & 70.20 & 91.21 & 89.02 & 92.39 & 89.32 & 88.35 & 82.79 & 80.96 & 78.11 \\

  Zhang et al.~\cite{Zhang2019DetectingAS} & CycleGAN & 55.39 & 100.0 & 75.08 & 55.11 & 66.08 & 100.0 & 45.18 & 47.46 & 57.12 & 53.61 & 50.98 & 57.72 & 77.72 & 77.25 & 76.47 & 68.58 & 64.58 & 61.92 & 67.77 & 66.21 \\

  Ojha et al.~\cite{ojha2023fakedetect} & CLIP-RN50 & 100.0 & 99.46 & 99.59 & 97.24 & 99.98 & 99.60 & 82.45 & 61.32 & 79.02 & 96.72 & 99.00 & 87.77 & 99.14 & 92.15 & 99.17 & 94.74 & 95.34 & 94.57 & 97.15 & 93.38 \\

  Tan et al.~\cite{Tan2024RethinkingTU} & S-ResNet50 & 100.0 & 90.49 & 84.55 & 99.61 & 77.76 & 100 & 90.15 & 91.21 & 87.34 & 67.59 & 67.72 & 94.81 & 98.39 & 96.94 & 98.56 & 88.65 & 92.66 & 91.22 & 93.58 & 90.06 \\
\hline
  \multirow{4}{*}{Wang et al.~\cite{wang2019cnngenerated}} & ResNet50$^{0.5}$ & 100.0 & 96.83 & 88.24 & 98.51 & 98.09 & 95.45 & 66.27 & 92.76 & 63.87 & 98.94 & 99.52 & 68.35 & 65.92 & 66.74 & 65.99 & 72.03 & 76.52 & 73.22 & 66.26 & 81.76 \\
   & VGG11 & 100.0 & 88.01 & 83.34 & 99.58 & 84.38 & 97.0 & 82.23 & 98.04 & 65.82 & 84.69 & 85.74 & 76.32 & 63.39 & 69.01 & 64.86 & 79.61 & 85.4 & 82.23 & 66.98 & 81.93 \\
   & ResNet50 & 100.0 & 90.39 & 83.17 & 97.69 & 93.16 & 96.82 & 76.79 & 91.52 & 74.54 & 95.2 & 96.04 & 76.57 & 79.13 & 74.86 & 80.8 & 81.64 & 85.18 & 81.46 & 62.09 & 85.11 \\
   & MobileNetv2 & 100.0 & 91.35 & 84.75 & 99.7 & 88.15 & 98.94 & 95.95 & 92.7 & 75.43 & 91.69 & 97.45 & 81.37 & 89.12 & 87.22 & 89.04 & 91.04 & 93.76 & 91.76 & 69.22 & 89.93 \\

  \multirow{4}{*}{Wang et al.~\cite{wang2019cnngenerated} + Ours} & ResNet50$^{0.5}$ & 100.0 & 92.97 & 90.06 & 98.15 & 97.80 & 87.94 & 73.09 & 89.92 & 74.57 & 96.97 & 98.00 & 70.98 & 77.75 & 72.44 & 78.21 & 78.53 & 82.61 & 79.27 & 78.16 & \textbf{85.13} \\
   & VGG11 & 100.0 & 92.4 & 92.62 & 98.33 & 91.4 & 99.13 & 91.35 & 95.3 & 77.66 & 96.54 & 95.62 & 81.33 & 80.73 & 79.77 & 80.11 & 83.86 & 86.95 & 85.88 & 79.0 & \textbf{88.84} \\
   & ResNet50 & 100.0 & 93.76 & 92.19 & 98.93 & 97.18 & 94.93 & 79.06 & 95.75 & 78.67 & 98.63 & 98.19 & 72.6 & 81.56 & 77.66 & 81.12 & 82.81 & 87.18 & 83.98 & 79.74 & \textbf{88.10} \\
   & MobileNetv2 & 100.0 & 92.1 & 86.09 & 96.41 & 83.76 & 99.99 & 93.85 & 90.81 & 84.0 & 88.84 & 89.52 & 85.97 & 93.69 & 92.91 & 93.32 & 90.79 & 93.67 & 92.15 & 81.9 & \textbf{91.04} \\
\hline
  Gragnaniello et al.~\cite{Gragnaniello2021} & ResNet50$^{nd}$ & 100.0 & 83.25 & 94.3 & 99.95 & 91.7 & 99.99 & 91.24 & 92.77 & 73.69 & 98.19 & 97.85 & 78.29 & 87.01 & 86.86 & 87.95 & 88.24 & 92.5 & 89.33 & 91.32 & 90.76 \\

  Gragnaniello et al.~\cite{Gragnaniello2021} + Ours & ResNet50$^{nd}$ & 100.0 & 94.07 & 97.61 & 99.92 & 98.54 & 99.99 & 94.0 & 95.46 & 81.47 & 96.72 & 95.47 & 80.14 & 94.35 & 94.24 & 94.77 & 90.32 & 93.46 & 91.05 & 96.29 & \textbf{94.10} \\
\hline
\end{tabular}}%
\vspace{1em}
\caption{Mean Average Precision (mAP) comparison with state-of-the-art methods across various deepfake detection benchmarks. \textbf{Bold} values indicate improvements of our approach over Wang et al.~\cite{wang2019cnngenerated} and Gragnaniello et al.~\cite{Gragnaniello2021} methods. S-ResNet50 denotes a shallow ResNet architecture, ResNet50$^{0.5}$ indicates data processing with 50\% probability of Gaussian blur and JPEG compression, and ResNet50$^{nd}$ means the first conv layer of ResNet was not downsampled.}
\label{tab_mapgen}
\end{table}

\begin{table*}[t]
\renewcommand{\arraystretch}{1.7}  % Adjust the row height
\centering
\Huge
\resizebox{\linewidth}{!}{
\begin{tabular}{cc|ccccccccccccccccccc|c}
\hline
\multirow{2}{*}{Method} & \multirow{2}{*}{Backbone} & \multicolumn{6}{c}{Generative Adversarial Networks} & \multirow{2}{*}{\parbox{1.5cm}{\centering  Deep\\Fake}} & \multicolumn{2}{c}{Low-Level Vision} & \multicolumn{2}{|c}{Perceptual Loss} & \multirow{2}{*}{\parbox{2.2cm}{\centering Guided}} & \multicolumn{3}{c|}{LDM} & \multicolumn{3}{c}{Glide} & \multirow{2}{*}{\parbox{2.8cm}{\centering DALL-E}} & average \\
\cline{3-8} \cline{10-13} \cline{15-20}
  &  & \parbox{1.5cm}{\centering Pro\\GAN} & \parbox{1.5cm}{\centering Cycle\\GAN} & \parbox{1.5cm}{\centering Big\\GAN} & \parbox{1.5cm}{\centering Style\\GAN} & \parbox{1.5cm}{\centering Gau\\GAN} & \parbox{1.5cm}{\centering Star\\GAN} &  & \parbox{1.5cm}{\centering SITD} & \parbox{1.5cm}{\centering SAN} & \parbox{1.5cm}{\centering CRN} & {\centering IMLE} &  & \parbox{1.5cm}{\centering 200\\steps} & \parbox{2.4cm}{\centering 200\\w/ CFG} & \parbox{1.5cm}{\centering 100\\steps} & \parbox{1.5cm}{\centering 100\\27} & \parbox{1.5cm}{\centering 50\\27} & \parbox{1.5cm}{\centering 100\\10} &  & AUROC \\
\hline
  \multirow{2}{*}{Chai et al.~\cite{DBLP:conf/eccv/ChaiBLI20}} & ResNet50 & 96.62 & 69.71 & 66.705 & 87.61 & 56.545 & 86.175 & 57.75 & 65.205 & 52.055 & 61.515 & 61.35 & 67.595 & 83.465 & 80.555 & 83.73 & 70.8 & 72.415 & 71.94 & 73.26 & 71.84 \\
  & Xception & 77.955 & 70.905 & 70.065 & 82.455 & 65.11 & 66.595 & 76.045 & 75.665 & 75.81 & 73.425 & 61.91 & 71.22 & 81.8 & 81.41 & 81.085 & 80.09 & 78.505 & 73.45 & 71.79 & 74.49 \\

  Natraj et al.~\cite{Nataraj2019DetectingGG} & - & 98.72 & 72.05 & 52.18 & 95.565 & 52.105 & 61.345 & 58.12 & 66.02 & 58.135 & 69.355 & 76.505 & 65.35 & 80.955 & 79.79 & 81.695 & 79.785 & 78.975 & 76.345 & 74.26 & 72.49 \\

  Zhang et al.~\cite{Zhang2019DetectingAS} & CycleGAN & 52.65 & 99.95 & 62.79 & 52.505 & 58.19 & 99.85 & 47.64 & 48.73 & 52.56 & 52.105 & 50.54 & 54.31 & 64.06 & 63.825 & 63.385 & 60.14 & 57.99 & 56.16 & 58.885 & 60.83 \\

  Ojha et al.~\cite{ojha2023fakedetect} & CLIP-RN50 & 100.0 & 98.98 & 97.045 & 89.62 & 99.74 & 98.30 & 74.525 & 62.16 & 68.26 & 78.11 & 85.50 & 78.90 & 96.665 & 82.955 & 96.765 & 86.905 & 87.595 & 86.355 & 91.965 & 87.38 \\

  Tan et al.~\cite{Tan2024RethinkingTU} & S-ResNet50 & 100.0 & 86.30 & 82.10 & 98.00 & 70.30 & 100 & 94.60 & 97.90 & 91.80 & 84.50 & 85.10 & 92.90 & 93.60 & 91.90 & 94.00 & 89.30 & 92.80 & 95.10 & 71.10 & 90.59 \\
\hline
  \multirow{4}{*}{Wang et al.~\cite{wang2019cnngenerated}} & ResNet50$^{0.5}$ & 100.0 & 93.1 & 85.7 & 98.9 & 94.8 & 96.4 & 85.0 & 97.2 & 75.1 & 98.0 & 98.9 & 77.5 & 75.0 & 72.1 & 75.6 & 83.8 & 87.1 & 85.3 & 59.9 & 85.77 \\
   & VGG11 & 100.0 & 89.50 & 86.10 & 99.50 & 87.00 & 96.10 & 82.00 & 97.70 & 68.10 & 90.20 & 91.20 & 76.80 & 64.30 & 66.80 & 64.50 & 78.10 & 83.80 & 81.00 & 62.90 & 82.40 \\
   & ResNet50 & 100.0 & 91.40 & 84.40 & 97.40 & 94.20 & 96.40 & 77.00 & 93.70 & 76.70 & 96.00 & 96.80 & 75.70 & 78.90 & 74.90 & 80.60 & 81.00 & 84.10 & 80.30 & 57.50 & 82.99 \\
   & MobileNetv2 & 100.0 & 92.40 & 87.30 & 99.70 & 89.60 & 98.70 & 96.00 & 92.40 & 75.20 & 92.20 & 97.90 & 81.00 & 88.40 & 85.90 & 88.00 & 90.10 & 92.90 & 91.00 & 65.50 & 89.12 \\

  \multirow{4}{*}{Wang et al.~\cite{wang2019cnngenerated} + Ours} & ResNet50$^{0.5}$ & 100.0 & 92.8 & 91.3 & 98.1 & 97.9 & 86.5 & 72.7 & 89.2 & 75.0 & 97.1 & 98.1 & 72.8 & 80.2 & 74.4 & 80.7 & 80.2 & 84.1 & 80.4 & 80.0 & \textbf{85.84} \\
   & VGG11 & 100.0 & 92.00 & 93.30 & 98.10 & 92.50 & 98.90 & 91.50 & 95.40 & 79.00 & 96.60 & 96.10 & 81.10 & 81.00 & 80.20 & 81.10 & 85.50 & 88.00 & 87.40 & 79.80 & \textbf{89.34} \\
   & ResNet50 & 100.0 & 93.20 & 92.90 & 98.90 & 97.20 & 94.30 & 80.50 & 95.50 & 75.60 & 98.70 & 98.40 & 74.00 & 83.00 & 79.40 & 83.10 & 84.80 & 88.60 & 85.60 & 82.00 & \textbf{88.72} \\
   & MobileNetv2 & 100.0 & 92.30 & 87.30 & 95.90 & 86.20 & 100 & 93.10 & 91.50 & 86.20 & 92.20 & 92.90 & 84.50 & 92.80 & 92.10 & 92.20 & 90.80 & 93.90 & 92.30 & 80.50 & \textbf{91.09} \\
\hline
  Gragnaniello et al.~\cite{Gragnaniello2021} & ResNet50$^{nd}$ & 100.0 & 87.20 & 95.50 & 99.90 & 94.00 & 100 & 89.20 & 94.70 & 74.30 & 98.90 & 98.50 & 79.80 & 87.30 & 86.70 & 88.20 & 87.90 & 92.20 & 89.30 & 91.80 & 91.33 \\

  Gragnaniello et al.~\cite{Gragnaniello2021} + Ours & ResNet50$^{nd}$ & 100.0 & 93.70 & 98.00 & 99.90 & 98.70 & 100 & 93.00 & 95.70 & 81.80 & 97.30 & 96.20 & 79.50 & 94.20 & 93.80 & 94.80 & 90.00 & 93.10 & 90.90 & 96.40 & \textbf{94.00} \\
\hline
\end{tabular}}
\vspace{1em}
\caption{Area Under Receiver Operating Characteristic (AUROC) comparison with state-of-the-art methods across various deepfake detection benchmarks. \textbf{Bold} values indicate improvements of our approach over Wang et al.~\cite{wang2019cnngenerated} and Gragnaniello et al.~\cite{Gragnaniello2021} methods. S-ResNet50 denotes a shallow ResNet architecture, ResNet50$^{0.5}$ indicates data processing with 50\% probability of Gaussian blur and JPEG compression, and ResNet50$^{nd}$ means the first conv layer of ResNet was not downsampled.}
\label{tab_aurocgen}
\end{table*}

As shown in Table \ref{tab_mapgen} and \ref{tab_aurocgen}, our approach (`+ Ours'), incorporating 15\% random masking across all frequency bands, consistently enhances performance when integrated with existing state-of-the-art (SOTA) methods. Specifically, the combination of our frequency-based masking technique with Wang et al.'s method~\cite{wang2019cnngenerated} results in improvements of mAP and AUROC for all backbones used. Here, S-ResNet50 refers to a shallow ResNet architecture, ResNet50$^{0.5}$ indicates data processing with 50\% probability of Gaussian blur and JPEG compression, and ResNet50$^{nd}$ denotes a modified ResNet where the first convolutional layer was not downsampled. Furthermore, integrating our method with that of Gragnaniello et al.~\cite{Gragnaniello2021} yields a performance that beats SOTA for all metrics. Notably, the gains are particularly pronounced in cross-dataset evaluations (non-ProGAN datasets), suggesting improved generalization capabilities. These results demonstrate the effectiveness of frequency masking, as it enables the model to learn more generalizable features for universal deepfake detection that are potentially overlooked by existing approaches. 

\subsection{Combined Augmentation Strategies}

\begin{figure}[t!]
\centering
\includegraphics[width=0.75\linewidth]{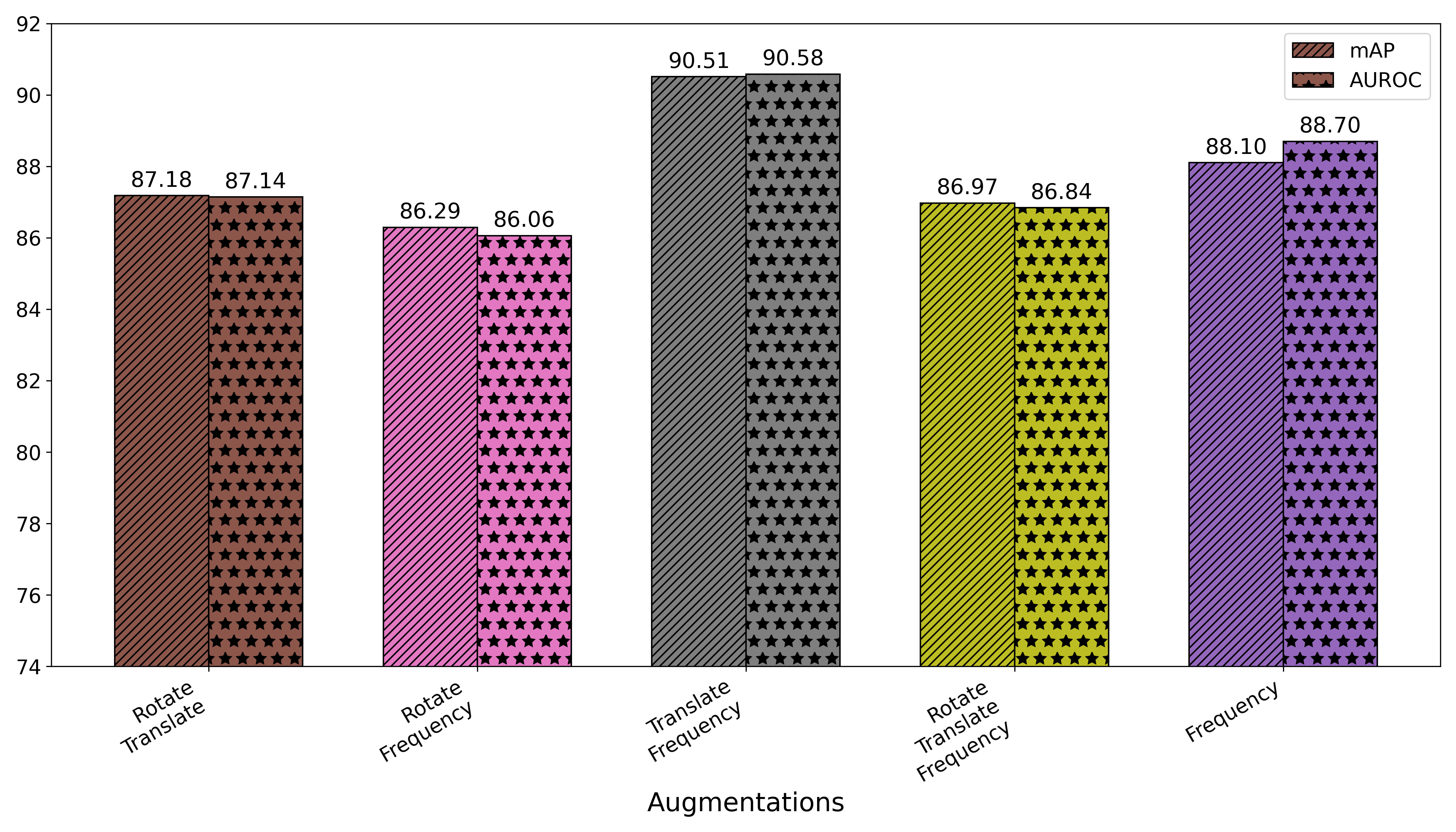}
\caption{Performance of combined augmentation strategies: Rotate+Translate, Rotate+Frequency, Translate+Frequency, Rotate+Translate+Frequency, and standalone Frequency masking. The Translate+Frequency combination achieves the highest performance, suggesting complementary benefits between spatial translation and spectral masking.}
\label{fig_combined_augs}
\end{figure}

To comprehensively evaluate the generalization advantage of frequency-domain masking, we compare it against combined geometric masking strategies (Figure~\ref{fig_combined_augs}). The combination of translation and rotation---a common robust training strategy---achieves 87.18\% mAP and 87.14\% AUROC, showing moderate improvement over individual geometric transformations but still falling short of standalone frequency masking (88.10\% mAP, 88.7\% AUROC). We believe this is because rotation is a stronger and more disruptive operation than translation: rotating faces changes their orientation and introduces interpolation artifacts, which may partly “wash out” the subtle traces that the detector needs to learn. In contrast, translation mostly shifts the content without heavily distorting local structures, which seems more helpful for robust training. 

Crucially, the integration of translation with frequency masking yields exceptional performance (90.51\% mAP, 90.58\% AUROC), suggesting that these two augmentations act in a complementary way: translation mainly breaks simple spatial patterns, while frequency masking targets inconsistencies in the spectral domain, so together they provide a richer set of examples. However, adding rotation to this combination (Rotate+Translate+Frequency) reduces performance to 86.97\% mAP and 86.84\% AUROC. A simple explanation is that stacking rotation on top of translation and frequency masking may over-perturb the images, making them less similar to realistic test data and encouraging the model to rely on artificial artifacts created by the augmentations rather than on genuine deepfake cues. Overall, these results confirm that carefully selected combined strategies can outperform individual approaches, with translation–frequency masking emerging as particularly effective for universal deepfake detection.

\subsection{Stability of Frequency Channel Selection}
\label{subsec:channel_stability}

\begin{table}[t!]
\centering
\begin{minipage}{0.48\textwidth}
\centering
\begin{tabular}{c|cc}
\hline
Masked Channel & mAP & AUROC \\ \hline
RGB & 88.10 & 88.72 \\
R & 87.01 & 87.58 \\
G & 89.85 & 90.12 \\
B & 87.17 & 88.02 \\ \hline
\end{tabular}
\caption{Color channel ablation}
\label{tab_ablation_color}
\end{minipage}
\hfill
\begin{minipage}{0.48\textwidth}
\centering
\begin{tabular}{c|cc}
\hline
Transform Type & mAP & AUROC \\ \hline
Fourier & 88.10 & 88.72 \\
Wavelet & 86.93 & 86.90 \\
Cosine & 88.52 & 89.00 \\ \hline
\end{tabular}
\caption{Frequency transform ablation}
\label{tab_ablation_transforms}
\end{minipage}
\end{table}
  % This includes both subtables

To assess the stability of our random frequency channel selection, we evaluate performance when masking specific color channels individually (Table~\ref{tab_ablation_color}). The table reports mAP/AUROC as follows: RGB 88.10/88.72, R 87.01/87.58, G 89.85/90.12, and B 87.17/88.02. Masking the green channel yields the highest score, while masking red or blue is close to the RGB baseline; overall, no single channel causes a substantial performance drop or gain. These results suggest that the method is not overly sensitive to which individual channel is masked, and masking all channels (RGB) remains a strong and competitive choice.

\subsection{Comparison of Frequency Transforms}
\label{subsec:transform_comparison}

We investigate alternative frequency transforms to validate our Fourier-based approach (Table~\ref{tab_ablation_transforms}). The Discrete Cosine Transform (DCT) slightly outperforms Fourier (88.52\% mAP, 89.00\% AUROC vs 88.10\% mAP, 88.72\% AUROC), potentially due to its energy compaction properties concentrating artifacts in fewer coefficients. Wavelet transforms underperform (86.93\% mAP, 86.90\% AUROC), possibly because their multi-resolution analysis dilutes generative artifacts across scales. Despite DCT's marginal gain, we retain Fourier transforms due to its established efficacy in prior forensic work~\cite{Corvi_2023_CVPR} and minimal performance difference.

\subsection{Augmentation Benefits across Pruning Levels}

\begin{table*}[t!]
\label{table:combined_sparsity}
\centering
\resizebox{0.98\linewidth}{!}{
\begin{tabular}{c}
\\
\hline
\textbf{MobileNetv2} \\
\resizebox{\linewidth}{!}{\tiny
\begin{tabular}{c|c|cc|cc|cc|cc}
\hline
\multirow{3}{*}{\parbox{1.8cm}{\centering Pruner}} & \multirow{3}{*}{\parbox{1.8cm}{\centering Augmentation}} & \multicolumn{8}{c}{\parbox{1.8cm}{\centering Structured Pruning }} \\
\cline{3-10}
\multicolumn{1}{c|}{} & \multicolumn{1}{c|}{} & \multicolumn{2}{c|}{\textit{p}=0.0} & \multicolumn{2}{c|}{\textit{p}=0.2} & \multicolumn{2}{c|}{\textit{p}=0.5} & \multicolumn{2}{c}{\textit{p}=0.8} \\
\cline{3-10}
\multicolumn{1}{c|}{} & \multicolumn{1}{c|}{} & mAP & AUROC & mAP & AUROC & mAP & AUROC & mAP & AUROC \\
\hline
\multirow{3}{*}{\parbox{1.5cm}{\centering Slim \cite{Liu2017LearningEC} \\ICCV'17}} 
    & Baseline~\cite{wang2019cnngenerated}                        & 89.93 & 89.12 & 76.03 & 75.15 & 53.95 & 53.47 & 66.65 & 64.65 \\
    & Translate                       & 90.05 & 89.81 & 74.90 & 73.94 & 61.10 & 60.88 & 68.38 & 66.72 \\
    & \textbf{Frequency (Ours)}       & \textbf{91.04} & \textbf{91.09} & \textbf{79.24} & \textbf{79.04} & \textbf{73.15} & \textbf{71.32} & \textbf{70.80} & \textbf{68.65} \\
\hline
\multirow{3}{*}{\parbox{1.5cm}{\centering LAMP \cite{Lee2020LayeradaptiveSF} \\ICLR'21}} 
    & Baseline~\cite{wang2019cnngenerated}                        & 89.93 & 89.12 & 80.63 & 80.85 & 58.80 & 59.02 & 63.36 & 63.96 \\
    & Translate                       & 90.05 & 89.81 & 79.62 & 80.17 & 57.73 & 58.64 & \textbf{68.80} & \textbf{68.04} \\
    & \textbf{Frequency (Ours)}       & \textbf{91.04} & \textbf{91.09} & \textbf{84.44} & \textbf{85.10} & \textbf{63.20} & \textbf{63.21} & 57.04 & 57.27 \\
\hline
\multirow{3}{*}{\parbox{1.5cm}{\centering GReg \cite{Wang2020NeuralPV} \\ICLR'21}} 
    & Baseline~\cite{wang2019cnngenerated}                        & 89.93 & 89.12 & 80.62 & 80.82 & 58.47 & 58.52 & 64.88 & 64.99 \\
    & Translate                       & 90.05 & 89.81 & 80.04 & 80.57 & 58.56 & 59.75 & \textbf{69.14} & \textbf{68.30} \\
    & \textbf{Frequency (Ours)}       & \textbf{91.04} & \textbf{91.09} & \textbf{84.44} & \textbf{85.10} & \textbf{62.05} & \textbf{62.51} & 58.20 & 57.96 \\
\hline
\multirow{3}{*}{\parbox{1.5cm}{\centering DepGraph \cite{Fang2023DepGraphTA} \\CVPR'23}} 
    & Baseline~\cite{wang2019cnngenerated}                        & 89.93 & 89.12 & 80.79 & 80.99 & 58.83 & 58.91 & 63.90 & 63.90 \\
    & Translate                       & 90.05 & 89.81 & 79.66 & 80.21 & 58.08 & 58.96 & \textbf{68.74} & \textbf{68.02} \\
    & \textbf{Frequency (Ours)}       & \textbf{91.04} & \textbf{91.09} & \textbf{84.67} & \textbf{85.33} & \textbf{63.39} & \textbf{63.57} & 56.87 & 57.09 \\
\hline
\multicolumn{2}{c|}{Parameters (M)} & \multicolumn{2}{c|}{2.225} & \multicolumn{2}{c|}{1.412} & \multicolumn{2}{c|}{0.580} & \multicolumn{2}{c}{0.092} \\
\multicolumn{2}{c|}{MACs (G)} & \multicolumn{2}{c|}{0.319} & \multicolumn{2}{c|}{0.193} & \multicolumn{2}{c|}{0.089} & \multicolumn{2}{c}{0.050} \\
\hline
\hline
\end{tabular}} \\
\\
\hline
\textbf{ResNet50} \\
\resizebox{\linewidth}{!}{\tiny
\begin{tabular}{c|c|cc|cc|cc|cc}
\hline
\multirow{3}{*}{\parbox{1.8cm}{\centering Pruner}} & \multirow{3}{*}{\parbox{1.8cm}{\centering Augmentation}} & \multicolumn{8}{c}{\parbox{1.8cm}{\centering Structured Pruning }} \\
\cline{3-10}
\multicolumn{1}{c|}{} & \multicolumn{1}{c|}{} & \multicolumn{2}{c|}{\textit{p}=0.0} & \multicolumn{2}{c|}{\textit{p}=0.2} & \multicolumn{2}{c|}{\textit{p}=0.5} & \multicolumn{2}{c}{\textit{p}=0.8} \\
\cline{3-10}
\multicolumn{1}{c|}{} & \multicolumn{1}{c|}{} & mAP & AUROC & mAP & AUROC & mAP & AUROC & mAP & AUROC \\
\hline
\multirow{3}{*}{\parbox{1.5cm}{\centering Slim \cite{Liu2017LearningEC} \\ICCV'17}} 
    & Baseline~\cite{wang2019cnngenerated}                        & 85.11 & 82.99 & 79.25 & 80.34 & 47.56 & 48.30 & \textbf{59.85} & \textbf{64.93} \\
    & Translate                       & 87.37 & 87.50 & \textbf{85.04} & 85.26 & \textbf{82.26} & \textbf{82.89} & 51.24 & 54.99 \\
    & \textbf{Frequency (Ours)}       & \textbf{88.10} & \textbf{88.70} & \underline{85.01} & \textbf{85.26} & \underline{69.77} & \underline{70.30} & \underline{55.17} & \underline{59.34} \\
\hline
\multirow{3}{*}{\parbox{1.5cm}{\centering LAMP \cite{Lee2020LayeradaptiveSF} \\ICLR'21}} 
    & Baseline~\cite{wang2019cnngenerated}                        & 85.11 & 82.99 & 82.39 & 82.47 & 71.08 & 72.82 & \textbf{64.03} & \textbf{65.69} \\
    & Translate                       & 87.37 & 87.50 & 84.13 & 84.14 & 77.14 & 78.74 & 61.87 & 64.00 \\
    & \textbf{Frequency (Ours)}       & \textbf{88.10} & \textbf{88.70} & \textbf{86.91} & \textbf{86.70} & \textbf{80.88} & \textbf{82.82} & 58.54 & 57.99 \\
\hline
\multirow{3}{*}{\parbox{1.5cm}{\centering GReg \cite{Wang2020NeuralPV} \\ICLR'21}} 
    & Baseline~\cite{wang2019cnngenerated}                        & 85.11 & 82.99 & 83.07 & 83.48 & 72.21 & 74.12 & \textbf{66.13} & \textbf{67.88} \\
    & Translate                       & 87.37 & 87.50 & 84.69 & 84.68 & 76.87 & 78.41 & 59.60 & 61.68 \\
    & \textbf{Frequency (Ours)}       & \textbf{88.10} & \textbf{88.70} & \textbf{86.85} & \textbf{86.84} & \textbf{80.51} & \textbf{82.46} & \underline{63.87} & \underline{65.98} \\
\hline
\multirow{3}{*}{\parbox{1.5cm}{\centering DepGraph \cite{Fang2023DepGraphTA} \\CVPR'23}} 
    & Baseline~\cite{wang2019cnngenerated}                        & 85.11 & 82.99 & 80.89 & 80.85 & 73.49 & 75.55 & \textbf{64.48} & \textbf{65.59} \\
    & Translate                       & 87.37 & 87.50 & 84.95 & 84.95 & 76.36 & 78.00 & 58.71 & 60.71 \\
    & \textbf{Frequency (Ours)}       & \textbf{88.10} & \textbf{88.70} & \textbf{85.06} & \textbf{85.56} & \textbf{80.55} & \textbf{82.18} & \underline{63.23} & \underline{64.75} \\
\hline
\multicolumn{2}{c|}{Parameters (M)} & \multicolumn{2}{c|}{23.51} & \multicolumn{2}{c|}{14.74} & \multicolumn{2}{c|}{5.894} & \multicolumn{2}{c}{0.858} \\
\multicolumn{2}{c|}{MACs (G)} & \multicolumn{2}{c|}{4.120} & \multicolumn{2}{c|}{2.514} & \multicolumn{2}{c|}{1.068} & \multicolumn{2}{c}{0.151} \\
\hline
\hline
\end{tabular}} \\
\end{tabular}}
\vspace{1em}
\caption{The performance of models trained using different augmentation techniques, pruned, and then finetuned is evaluated under various \textit{p} rates of structural pruning (i.e., 0.0, 0.2, 0.5, and 0.8). Top: MobileNetv2 results. Bottom: ResNet50 results. \textbf{Bold} indicates the best result, while an \underline{underline} represents the second-best result.}
\label{tab_pruning}
\end{table*}

To evaluate the effectiveness of our proposed frequency masking augmentation under model compression, we apply structured pruning at varying sparsity levels (\textit{p} = 0.0, 0.2, 0.5, 0.8) following with fine-tuning phase. The goal is to assess whether robustness introduced via augmentation can persist under severe parameter reduction.

We adopt several pruning strategies including Slim~\cite{Liu2017LearningEC}, LAMP~\cite{Lee2020LayeradaptiveSF}, GReg~\cite{Wang2020NeuralPV}, and DepGraph~\cite{Fang2023DepGraphTA}. The augmentation baseline refers to training without any masking or augmentation, following the protocol of Wang et al.~\cite{wang2019cnngenerated}. From Table~\ref{tab_pruning}, we observe consistent trends across both MobileNetv2 and ResNet50. Notably, when no pruning is applied (\( p=0.0 \)), frequency-based augmentation achieves the highest performance, validating its regularization strength during full-capacity training. 

As the pruning rate increases to moderate levels (\( p=0.2 \) and \( p=0.5 \)), frequency augmentation maintains its advantage across most pruning methods. Interestingly, translation-based augmentation often performs competitively, in some cases, emerging as the second best-performing method under severe pruning. This observation aligns with earlier results in Figure~\ref{fig_masking_types}, where frequency and translation augmentations consistently outperformed others.

However, a failure case is observed when the pruning rate reaches \( p=0.8 \). At this extreme compression level, performance degrades significantly for all augmentations. Frequency masking, in particular, shows a sharp drop in several configurations (e.g., MobileNetv2 with DepGraph), suggesting that aggressive sparsity nullifies the regularization benefit from masking.

Overall, these results highlight the importance of selecting augmentation strategies that not only improve generalization but also synergize with pruning to retain performance in resource-constrained settings.

\subsection{Use Case: Synthetic Dataset from Aquaculture}

\begin{table}[t]
\renewcommand{\arraystretch}{1.7}  % Adjust the row height
\centering
\tiny
\resizebox{0.75\linewidth}{!}{
\begin{tabular}{c|cc|cc}
\hline
\multirow{2}{*}{Method} & \multicolumn{2}{c|}{ControlNet~\cite{Zhang2023AddingCC}} & \multicolumn{2}{c}{Stable Diffusion~\cite{Rombach2021HighResolutionIS}} \\
\cline{2-5}
  & mAP & AUROC & mAP & AUROC \\
\hline
    Tan et al.~\cite{Tan2024RethinkingTU} & 54.67 & 43.81 & 50.47 & 35.76 \\
    Gragnaniello et al.~\cite{Gragnaniello2021} & 63.16 & 56.86 & 61.17 & 57.31 \\
    % Gragnaniello et al.~\cite{Gragnaniello2021} + \textit{Ours} & 69.37 & 65.41 & 72.93 & 69.22 \\
    Wang et al.~\cite{wang2019cnngenerated} & 73.38 & 71.46 & 68.69 & 68.77 \\
    Ours & \textbf{79.84} & \textbf{80.39} & \textbf{84.93} & \textbf{85.92} \\
\hline
\end{tabular}}
\vspace{1em}
\caption{mAP and AUROC results on the FakeFish dataset generated using ControlNet~\cite{Zhang2023AddingCC} and Stable Diffusion~\cite{Rombach2021HighResolutionIS}.}
\label{tab_fakefish}
\end{table}

Deepfake detection methods can be extended beyond general-purpose datasets to more specialized domains, such as aquaculture. In this field, maintaining the quality and health of fish stocks is critical for both food safety and economic profit. As collecting large-scale, high-quality real data is difficult and expensive, synthetic datasets are often used to augment training. However, these generated images are not always realistic or high quality, and may not be reliable for downstream tasks such as disease detection or quality control. In extreme cases, poor-quality synthetic images may be misused to fake healthy appearances of fish, potentially masking signs of disease.

To explore this idea, we collected a synthetic dataset called \textit{FakeFish}, generated using two popular image synthesis methods: ControlNet~\cite{Zhang2023AddingCC} and Stable Diffusion~\cite{Rombach2021HighResolutionIS}. Table~\ref{tab_fakefish} shows our model outperforming prior work by a large margin in both mean Average Precision (mAP) and Area Under the ROC Curve (AUROC). This suggests that our approach is effective at identifying low-quality or manipulated synthetic images in aquaculture scenarios.

\begin{figure}[htbp]
    \centering
    
    % First row: Real images
    \begin{subfigure}[b]{0.23\textwidth}
        \centering
        \includegraphics[width=\textwidth]{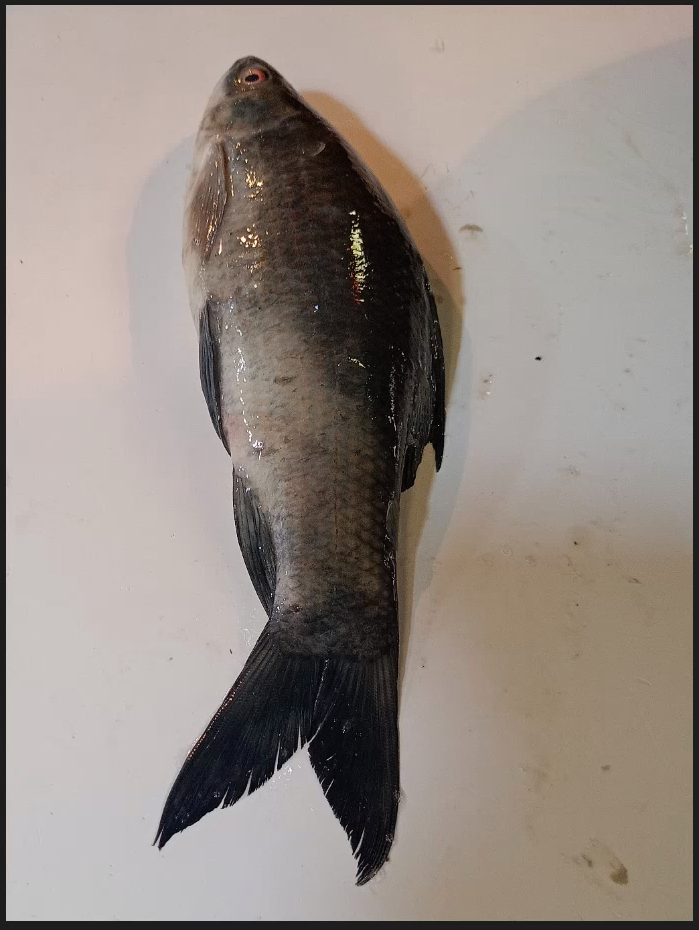}
    \end{subfigure}
    \hfill
    \begin{subfigure}[b]{0.23\textwidth}
        \centering
        \includegraphics[width=\textwidth]{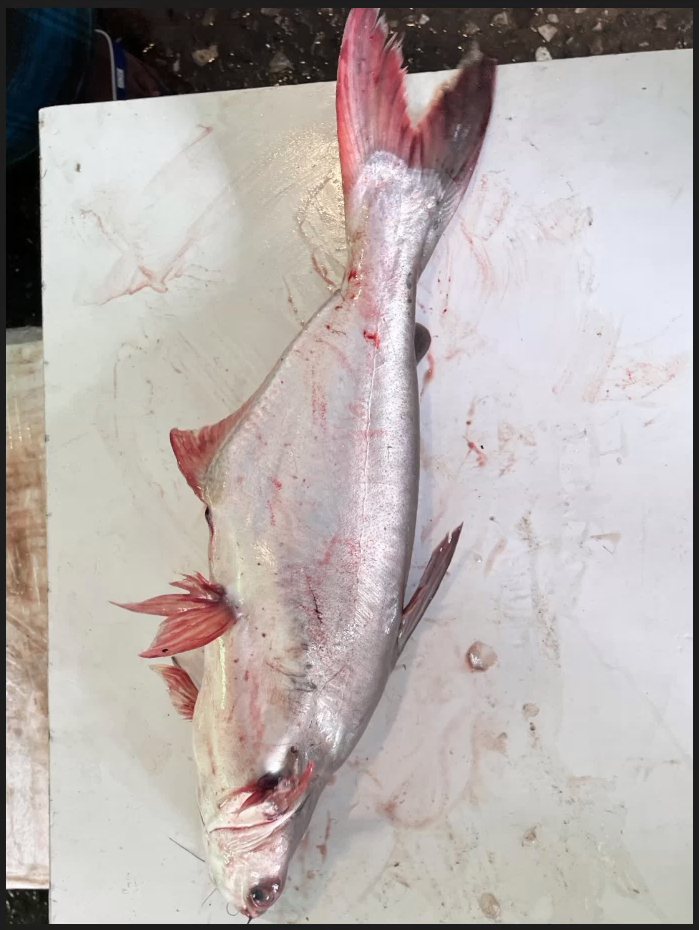}
    \end{subfigure}
    \hfill
    \begin{subfigure}[b]{0.23\textwidth}
        \centering
        \includegraphics[width=\textwidth]{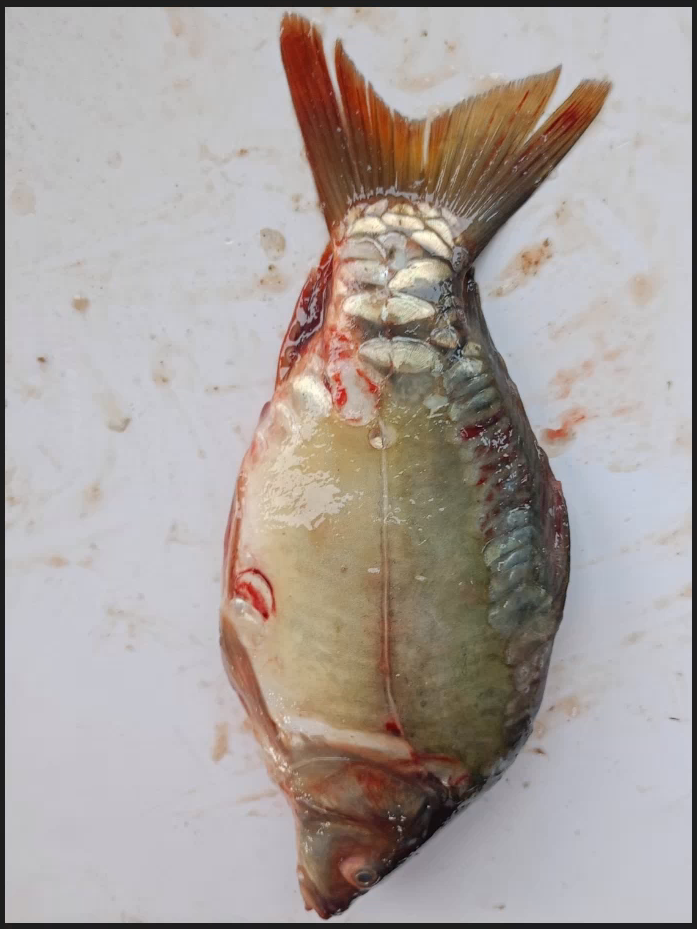}
    \end{subfigure}
    \hfill
    \begin{subfigure}[b]{0.23\textwidth}
        \centering
        \includegraphics[width=\textwidth]{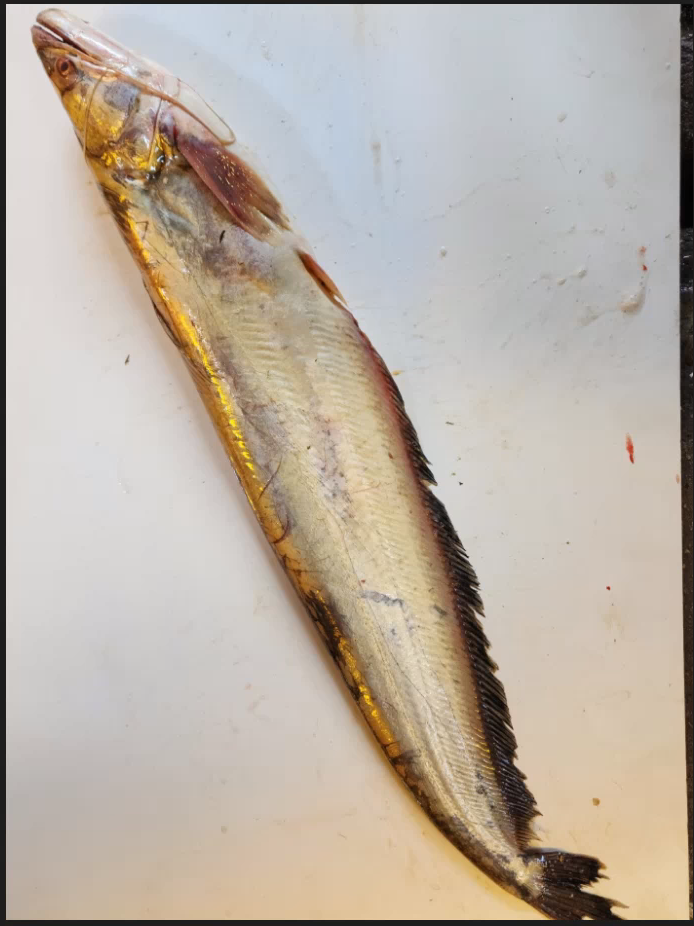}
    \end{subfigure}
    \caption*{Real}
    
    \vspace{0.2cm} % Add some vertical space between rows
    
    % Second row: ControlNet images
    \begin{subfigure}[b]{0.23\textwidth}
        \centering
        \includegraphics[width=\textwidth]{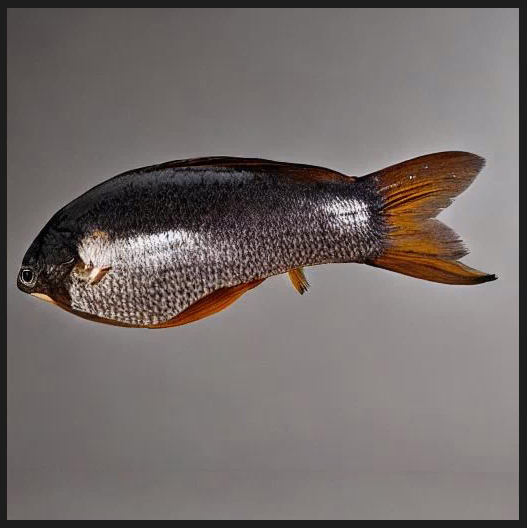}
    \end{subfigure}
    \hfill
    \begin{subfigure}[b]{0.23\textwidth}
        \centering
        \includegraphics[width=\textwidth]{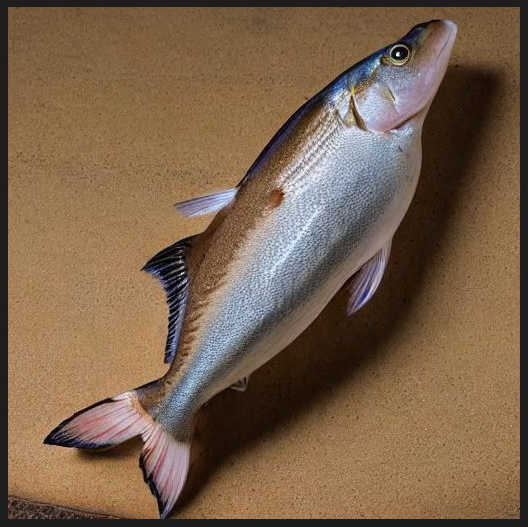}
    \end{subfigure}
    \hfill
    \begin{subfigure}[b]{0.23\textwidth}
        \centering
        \includegraphics[width=\textwidth]{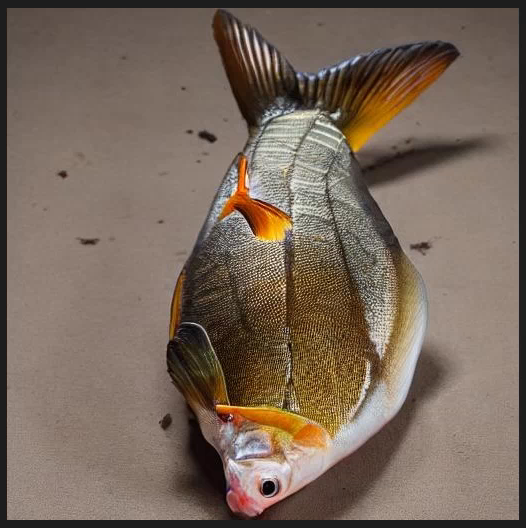}
    \end{subfigure}
    \hfill
    \begin{subfigure}[b]{0.23\textwidth}
        \centering
        \includegraphics[width=\textwidth]{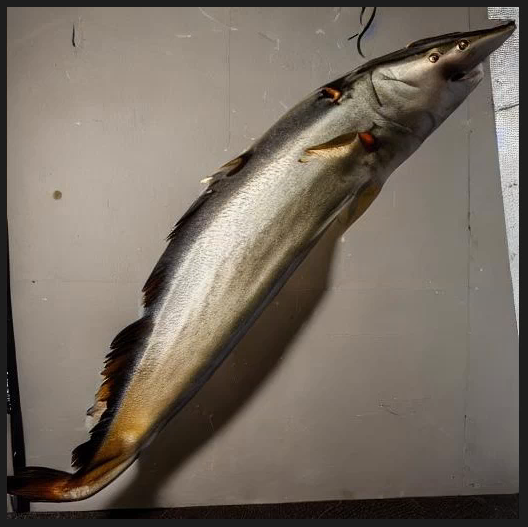}
    \end{subfigure}
    \caption*{ControlNet}
    
    \vspace{0.2cm} % Add some vertical space between rows
    
    % Third row: Stable Diffusion images
    \begin{subfigure}[b]{0.23\textwidth}
        \centering
        \includegraphics[width=\textwidth]{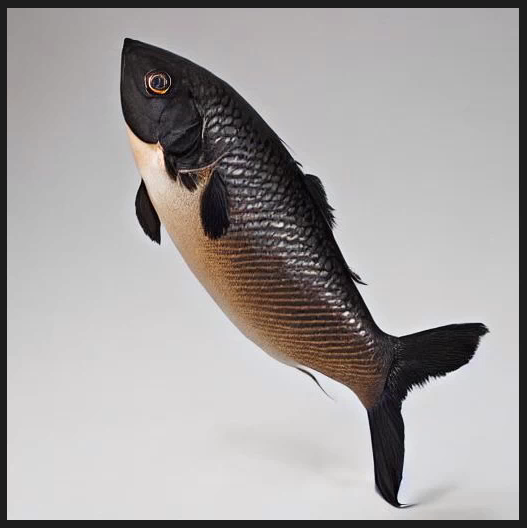}
    \end{subfigure}
    \hfill
    \begin{subfigure}[b]{0.23\textwidth}
        \centering
        \includegraphics[width=\textwidth]{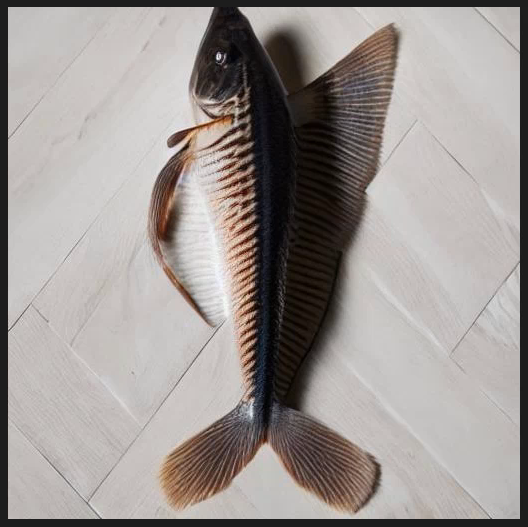}
    \end{subfigure}
    \hfill
    \begin{subfigure}[b]{0.23\textwidth}
        \centering
        \includegraphics[width=\textwidth]{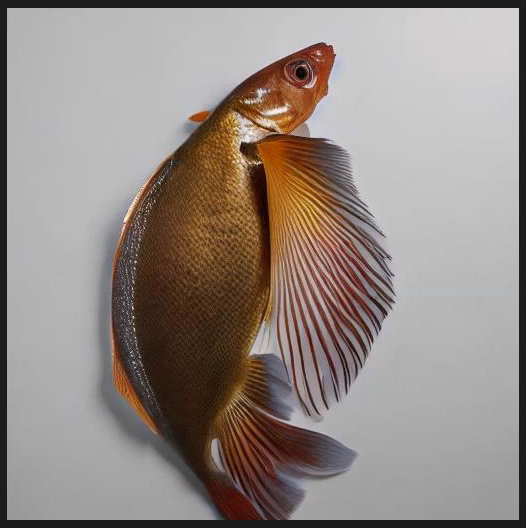}
    \end{subfigure}
    \hfill
    \begin{subfigure}[b]{0.23\textwidth}
        \centering
        \includegraphics[width=\textwidth]{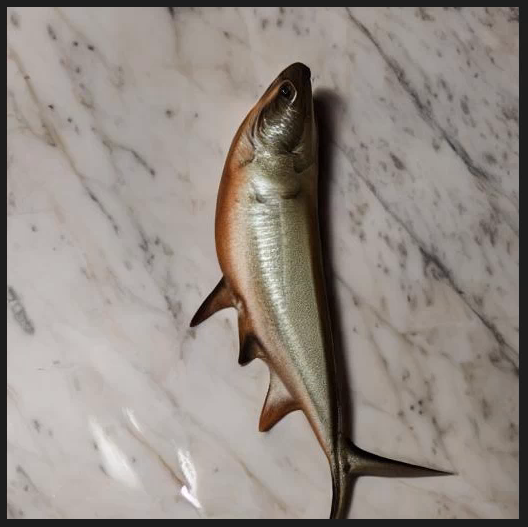}
    \end{subfigure}
    \caption*{Stable Diffusion}
    
    \caption{Visualization of Real and AI-generated fish images from ControlNet~\cite{Zhang2023AddingCC} and Stable Diffusion~\cite{Rombach2021HighResolutionIS} in the FakeFish dataset.}
    \label{fig_visfake}
\end{figure}

Figure~\ref{fig_visfake} shows visual examples of the generated images used in the FakeFish dataset, highlighting different levels of realism and artifacts across generation methods.

\section{Conclusion and Future Work}

This work addressed a fundamental challenge in deepfake detection: improving generalization across diverse synthetic media while maintaining computational efficiency. Our experiments demonstrate that our proposed frequency-domain masking during training offers a viable solution, enabling detectors to retain discriminative features for deepfake detection even when the number of parameters is severely reduced. This finding directly responds to the growing need for sustainable and generalizable deepfake detection systems.

From a theoretical view, frequency masking can generalize well because it removes easy, generator-specific shortcuts while keeping the overall layout of the image. Many generators leave narrow or repeated patterns in certain frequency bands that a model can memorize. By randomly dropping parts of the spectrum during training, the model cannot rely on those brittle cues and is pushed to use signals that stay stable across datasets, such as shapes, edges at different scales, and color consistency. In Fourier terms, training sees many spectrally corrupted versions of the same image; the loss effectively averages over these views, like data augmentation or dropout in the frequency domain, which makes the decision surface smoother and less sensitive to small spectral shifts. Because we keep phase and mask amplitude, the spatial structure remains while fine textures tied to specific synthesis pipelines are weakened, so the network learns features tied to structure rather than artifacts. These produces features that transfer better across generators and also explain why the gains survive pruning: the useful signal is spread across bands and is less redundant, so it persists even when capacity is reduced.

In summary, frequency masking consistently outperforms spatial and geometric transformations, with an optimal masking ratio around 15\%. Translation combined with frequency masking is complementary and yields the strongest scores, whereas rotation+translation remains below frequency alone and adding rotation to translation+frequency can hurt. Performance is largely insensitive to which color channel is masked. While DCT is slightly stronger than FFT and DWT underperforms, FFT remains a simple, competitive default. These gains persist under model pruning, indicating features that transfer and survive compression, and we also observe strong real-world performance in aquaculture. At the same time, performance drops under extreme pruning (80\%), the approach still needs substantial finetuning data for some domains, and our evaluation focuses on images rather than video, where temporal frequency cues may matter.

\subsection{Suggestions for Future Work}
The framework could be extended to video by incorporating temporal frequency analysis. There's also significant potential in creating adaptive masking strategies that automatically adjust to different generator types and select transforms (e.g., FFT vs. DCT) based on data characteristics. These extensions would build upon the current findings while addressing the identified limitations, moving toward more comprehensive deepfake detection solutions.
\begin{acks}
This research is supported by the National Research Foundation, Singapore under its AI Singapore Programmes (AISG Award No.: AISG2-TC-2022-007); The Agency for Science, Technology and Research (A*STAR) under its MTC Programmatic Funds (Grant No. M23L7b0021). This research is supported by the National Research Foundation, Singapore and Infocomm Media Development Authority under its Trust Tech Funding Initiative. Any opinions, findings and conclusions or recommendations expressed in this material are those of the author(s) and do not reflect the views of National Research Foundation, Singapore and Infocomm Media Development Authority.

The computational work was supported by resources from Sigma2—the National Infrastructure for High-Performance Computing and Data Storage in Norway—including access to the LUMI supercomputer (project code NN11074K). The LUMI supercomputer is owned by the EuroHPC Joint Undertaking and is hosted by CSC (Finland) and the LUMI consortium through Sigma2—Norway. Supplementary computational support was provided by the Orion High Performance Computing Center (OHPCC) at the Norwegian University of Life Sciences (NMBU).

The first author is supported by a PhD scholarship from the Faculty of Science and Technology (REALTEK), Norwegian University of Life Sciences (NMBU). We also thank the editors and reviewers for their insightful comments and constructive feedback.
\end{acks}

\bibliographystyle{ACM-Reference-Format}
\bibliography{main}

@String{Computing = "Computing" }

@String{Computer = "{IEEE} Computer" }

@String{Springer = "Springer-Verlag" }

@InProceedings{Chandrasegaran_2022_ECCV,
  author       = {K. Chandrasegaran and
                  N. Tran and
                  A. Binder and
                  N. Cheung},
  title        = {Discovering Transferable Forensic Features for CNN-Generated Images
                  Detection},
  booktitle    = {Computer Vision - {ECCV} 2022 - 17th European Conference, Tel Aviv,
                  Israel, October 23-27, 2022, Proceedings, Part {XV}},
  series       = {Lecture Notes in Computer Science},
  volume       = {13675},
  pages        = {671--689},
  publisher    = {Springer},
  year         = {2022},
  url          = {https://doi.org/10.1007/978-3-031-19784-0\_39},
  doi          = {10.1007/978-3-031-19784-0\_39},
  bibsource    = {dblp computer science bibliography, https://dblp.org}
}

@inproceedings{wang2019cnngenerated,
  author       = {S. Wang and
                  O. Wang and
                  R. Zhang and
                  A. Owens and
                  A. A. Efros},
  title        = {CNN-Generated Images Are Surprisingly Easy to Spot... for Now},
  booktitle    = {2020 {IEEE/CVF} Conference on Computer Vision and Pattern Recognition,
                  {CVPR} 2020, Seattle, WA, USA, June 13-19, 2020},
  pages        = {8692--8701},
  publisher    = {Computer Vision Foundation / {IEEE}},
  year         = {2020},
  url          = {https://openaccess.thecvf.com/content\_CVPR\_2020/html/Wang\_CNN-Generated\_Images\_Are\_Surprisingly\_Easy\_to\_Spot...\_for\_Now\_CVPR\_2020\_paper.html},
  doi          = {10.1109/CVPR42600.2020.00872},
  bibsource    = {dblp computer science bibliography, https://dblp.org}
}

@InProceedings{Corvi_2023_CVPR,
  author       = {R. Corvi and
                  D. Cozzolino and
                  G. Poggi and
                  K. Nagano and
                  L. Verdoliva},
  title        = {Intriguing properties of synthetic images: from generative adversarial
                  networks to diffusion models},
  booktitle    = {{IEEE/CVF} Conference on Computer Vision and Pattern Recognition,
                  {CVPR} 2023 - Workshops, Vancouver, BC, Canada, June 17-24, 2023},
  pages        = {973--982},
  publisher    = {{IEEE}},
  year         = {2023},
  url          = {https://doi.org/10.1109/CVPRW59228.2023.00104},
  doi          = {10.1109/CVPRW59228.2023.00104},
  bibsource    = {dblp computer science bibliography, https://dblp.org}
}

@InProceedings{Corvi_2023_ICASSP,
  author       = {R. Corvi and
                  D. Cozzolino and
                  G. Zingarini and
                  G. Poggi and
                  K. Nagano and
                  L. Verdoliva},
  title        = {On the detection of synthetic images generated by diffusion models},
  booktitle      = {IEEE International Conference on Acoustics, Speech and Signal Processing (ICASSP)},
  volume       = {abs/2211.00680},
  year         = {2023},
  url          = {https://doi.org/10.48550/arXiv.2211.00680},
  doi          = {10.48550/arXiv.2211.00680},
  eprinttype    = {arXiv},
  eprint       = {2211.00680},
  bibsource    = {dblp computer science bibliography, https://dblp.org}
}

@inproceedings{Gragnaniello2021,
  author       = {D. Gragnaniello and
                  D. Cozzolino and
                  F. Marra and
                  G. Poggi and
                  L. Verdoliva},
  title        = {Are {GAN} Generated Images Easy to Detect? {A} Critical Analysis of
                  the State-Of-The-Art},
  booktitle    = {2021 {IEEE} International Conference on Multimedia and Expo, {ICME}
                  2021, Shenzhen, China, July 5-9, 2021},
  pages        = {1--6},
  publisher    = {{IEEE}},
  year         = {2021},
  url          = {https://doi.org/10.1109/ICME51207.2021.9428429},
  doi          = {10.1109/ICME51207.2021.9428429},
  bibsource    = {dblp computer science bibliography, https://dblp.org}
}

@inproceedings{ojha2023fakedetect,
      title={Towards Universal Fake Image Detectors that Generalize Across Generative Models}, 
      author={Ojha, U. and Li, Y. and Lee, Y.},
      booktitle={2023 IEEE/CVF Conference on Computer Vision and Pattern Recognition (CVPR)},
      year={2023},
      pages={24480-24489},
}

@inproceedings{DBLP:conf/eccv/ChaiBLI20,
  author       = {L. Chai and
                  D. Bau and
                  S. Lim and
                  P. Isola},
  title        = {What Makes Fake Images Detectable? Understanding Properties that Generalize},
  booktitle    = {Computer Vision - {ECCV} 2020 - 16th European Conference, Glasgow,
                  UK, August 23-28, 2020, Proceedings, Part {XXVI}},
  series       = {Lecture Notes in Computer Science},
  volume       = {12371},
  pages        = {103--120},
  publisher    = {Springer},
  year         = {2020},
  doi          = {10.1007/978-3-030-58574-7\_7},
}

@inproceedings{Chen2022OSTIG,
  title={OST: Improving Generalization of DeepFake Detection via One-Shot Test-Time Training},
  author={Liang Chen and Yong Zhang and Yibing Song and Jue Wang and Lingqiao Liu},
  booktitle={Neural Information Processing Systems},
  year={2022},
  url={https://api.semanticscholar.org/CorpusID:258509220}
}

@article{DBLP:journals/corr/abs-2302-02615,
  title={Rethinking Out-of-distribution (OOD) Detection: Masked Image Modeling is All You Need},
  author={J. Li and P. Chen and S. Yu and Z. He and S. Liu and J. Jia},
  journal={2023 IEEE/CVF Conference on Computer Vision and Pattern Recognition (CVPR)},
  year={2023},
  pages={11578-11589},
  url={https://api.semanticscholar.org/CorpusID:256615563}
}

@inproceedings{DBLP:conf/iclr/0002LZ0OL23,
  author       = {J. Xie and
                  W. Li and
                  X. Zhan and
                  Z. Liu and
                  Y. Ong and
                  C. Loy},
  title        = {Masked Frequency Modeling for Self-Supervised Visual Pre-Training},
  booktitle    = {The Eleventh International Conference on Learning Representations,
                  {ICLR} 2023, Kigali, Rwanda, May 1-5, 2023},
  publisher    = {OpenReview.net},
  year         = {2023},
  url          = {https://openreview.net/pdf?id=9-umxtNPx5E},
  bibsource    = {dblp computer science bibliography, https://dblp.org}
}

@inproceedings{DBLP:conf/nips/HuangCGXZLLX22,
  author       = {J. Huang and
                  K. Cui and
                  D. Guan and
                  A. Xiao and
                  F. Zhan and
                  S. Lu and
                  S. Liao and
                  E. Xing},
  title        = {Masked Generative Adversarial Networks are Data-Efficient Generation
                  Learners},
  booktitle    = {Neural Information Processing Systems},
  year         = {2022},
  url          = {http://papers.nips.cc/paper\_files/paper/2022/hash/0efcb1885b8534109f95ca82a5319d25-Abstract-Conference.html},
  bibsource    = {dblp computer science bibliography, https://dblp.org}
}

@inproceedings{DBLP:conf/cvpr/HeCXLDG22,
  author       = {K. He and
                  X. Chen and
                  S. Xie and
                  Y. Li and
                  P. Doll{\'{a}}r and
                  R. B. Girshick},
  title        = {Masked Autoencoders Are Scalable Vision Learners},
  booktitle    = {{IEEE/CVF} Conference on Computer Vision and Pattern Recognition,
                  {CVPR} 2022, New Orleans, LA, USA, June 18-24, 2022},
  pages        = {15979--15988},
  publisher    = {{IEEE}},
  year         = {2022},
  url          = {https://doi.org/10.1109/CVPR52688.2022.01553},
  doi          = {10.1109/CVPR52688.2022.01553},
  bibsource    = {dblp computer science bibliography, https://dblp.org}
}

@article{mirsky2020deepfake,
  title={The Creation and Detection of Deepfakes},
  author={Yisroel Mirsky and Wenke Lee},
  journal={ACM Computing Surveys (CSUR)},
  year={2020},
  volume={54},
  pages={1 - 41},
  url={https://api.semanticscholar.org/CorpusID:216080410}
}

@article{rombach2021diffusion,
  title={High-Resolution Image Synthesis with Latent Diffusion Models},
  author={Robin Rombach and A. Blattmann and Dominik Lorenz and Patrick Esser and Bj{\"o}rn Ommer},
  journal={2022 IEEE/CVF Conference on Computer Vision and Pattern Recognition (CVPR)},
  year={2021},
  pages={10674-10685},
  url={https://api.semanticscholar.org/CorpusID:245335280}
}

@article{Abdollahzadeh2023ASO,
  title={A Survey on Generative Modeling with Limited Data, Few Shots, and Zero Shot},
  author={Milad Abdollahzadeh and Touba Malekzadeh and Christopher T. H. Teo and Keshigeyan Chandrasegaran and Guimeng Liu and Ngai-Man Cheung},
  journal={ArXiv},
  year={2023},
  volume={abs/2307.14397},
  url={https://api.semanticscholar.org/CorpusID:260202878}
}

@article{Tan2024FrequencyAwareDD,
  title={Frequency-Aware Deepfake Detection: Improving Generalizability through Frequency Space Learning},
  author={Chuangchuang Tan and Yao Zhao and Shikui Wei and Guanghua Gu and Ping Liu and Yunchao Wei},
  journal={AAAI},
  year={2024},
  volume={abs/2403.07240},
  url={https://api.semanticscholar.org/CorpusID:268890333}
}

@article{Coccomini2024DeepfakeDW,
  title={Deepfake Detection without Deepfakes: Generalization via Synthetic Frequency Patterns Injection},
  author={Davide Alessandro Coccomini and Roberto Caldelli and Claudio Gennaro and Giuseppe Fiameni and Giuseppe Amato and Fabrizio Falchi},
  journal={ArXiv},
  year={2024},
  volume={abs/2403.13479},
  url={https://api.semanticscholar.org/CorpusID:268537294}
}

@article{Zhang2019DetectingAS,
  title={Detecting and Simulating Artifacts in GAN Fake Images},
  author={Xu Zhang and Svebor Karaman and Shih-Fu Chang},
  journal={2019 IEEE International Workshop on Information Forensics and Security (WIFS)},
  year={2019},
  pages={1-6},
  url={https://api.semanticscholar.org/CorpusID:196622700}
}

@article{Nataraj2019DetectingGG,
  title={Detecting GAN generated Fake Images using Co-occurrence Matrices},
  author={Lakshmanan Nataraj and Tajuddin Manhar Mohammed and B. S. Manjunath and Shivkumar Chandrasekaran and Arjuna Flenner and Jawadul H. Bappy and Amit K. Roy-Chowdhury},
  journal={ArXiv},
  year={2019},
  volume={abs/1903.06836},
  url={https://api.semanticscholar.org/CorpusID:81982547}
}

@article{Qian2020ThinkingIF,
  title={Thinking in Frequency: Face Forgery Detection by Mining Frequency-aware Clues},
  author={Yuyang Qian and Guojun Yin and Lu Sheng and Zixuan Chen and Jing Shao},
  journal={ECCV},
  year={2020},
}

@article{Frank2020LeveragingFA,
  title={Leveraging Frequency Analysis for Deep Fake Image Recognition},
  author={Joel Cameron Frank and Thorsten Eisenhofer and Lea Sch{\"o}nherr and Asja Fischer and Dorothea Kolossa and Thorsten Holz},
  journal={ICML},
  year={2020},
}

@inproceedings{Le2021ADDFA,
  title={ADD: Frequency Attention and Multi-View based Knowledge Distillation to Detect Low-Quality Compressed Deepfake Images},
  author={Binh Minh Le and Simon S. Woo},
  booktitle={AAAI Conference on Artificial Intelligence},
  year={2021},
  url={https://api.semanticscholar.org/CorpusID:244920788}
}

@article{Luo2021GeneralizingFF,
  title={Generalizing Face Forgery Detection with High-frequency Features},
  author={Yucheng Luo and Yong Zhang and Junchi Yan and Wei Liu},
  journal={2021 IEEE/CVF Conference on Computer Vision and Pattern Recognition (CVPR)},
  year={2021},
}

@inproceedings{Jeong2022FrePGANRD,
  title={FrePGAN: Robust Deepfake Detection Using Frequency-level Perturbations},
  author={Yonghyun Jeong and Doyeon Kim and Youngmin Ro and Jongwon Choi},
  booktitle={AAAI Conference on Artificial Intelligence},
  year={2022},
  url={https://api.semanticscholar.org/CorpusID:246634415}
}

@article{Li2024FreqBlenderED,
  title={FreqBlender: Enhancing DeepFake Detection by Blending Frequency Knowledge},
  author={Hanzhe Li and Yuezun Li and Jiaran Zhou and Bin Li and Junyu Dong},
  journal={Neural Information Processing Systems},
  year={2024},
  volume={abs/2404.13872},
  url={https://api.semanticscholar.org/CorpusID:269293651}
}

@article{Tian2023FrequencyAwareAF,
  title={Frequency-Aware Attentional Feature Fusion for Deepfake Detection},
  author={Cheng Tian and Zhiming Luo and Guimin Shi and Shaozi Li},
  journal={ICASSP 2023 - 2023 IEEE International Conference on Acoustics, Speech and Signal Processing (ICASSP)},
  year={2023},
  pages={1-5},
  url={https://api.semanticscholar.org/CorpusID:258533739}
}

@article{Tan2024RethinkingTU,
  title={Rethinking the Up-Sampling Operations in CNN-Based Generative Network for Generalizable Deepfake Detection},
  author={Chuangchuang Tan and Huan Liu and Yao Zhao and Shikui Wei and Guanghua Gu and Ping Liu and Yunchao Wei},
  journal={2024 IEEE/CVF Conference on Computer Vision and Pattern Recognition (CVPR)},
  year={2024},
  pages={28130-28139},
  url={https://api.semanticscholar.org/CorpusID:266348433}
}

@article{Chen2024MaskedCD,
  title={Masked Conditional Diffusion Model for Enhancing Deepfake Detection},
  author={Tiewen Chen and Shanmin Yang and Shu Hu and Zhenghan Fang and Ying Fu and Xi Wu and Xin Wang},
  journal={2024 International Joint Conference on Neural Networks (IJCNN)},
  year={2024},
  pages={1-7},
  url={https://api.semanticscholar.org/CorpusID:267364978}
}

@article{Das2023UnmaskingDM,
  title={Unmasking Deepfakes: Masked Autoencoding Spatiotemporal Transformers for Enhanced Video Forgery Detection},
  author={Sayantan Das and Mojtaba Kolahdouzi and Levent {\"O}zparlak and Will Hickie and Ali Etemad},
  journal={2023 IEEE International Joint Conference on Biometrics (IJCB)},
  year={2023},
  pages={1-11},
  url={https://api.semanticscholar.org/CorpusID:259841109}
}

@article{Chen2022DefakeHopAE,
  title={DefakeHop++: An Enhanced Lightweight Deepfake Detector},
  author={Hong-Shuo Chen and Shuowen Hu and Suya You and C.-C. Jay Kuo},
  journal={APSIPA Transactions on Signal and Information Processing},
  year={2022},
  volume={abs/2205.00211},
  url={https://api.semanticscholar.org/CorpusID:248496794}
}

@article{K2023CompressedDD,
  title={Compressed Deepfake Detection using Spatio-Temporal Approach with Model Pruning},
  author={Vidya K and Praveen Ramesh and Hrithik Viknesh and Sanjay Devanand},
  journal={Procedia Computer Science},
  year={2023},
  url={https://api.semanticscholar.org/CorpusID:266828037}
}

@article{Lim2024DistilDIREAS,
  title={DistilDIRE: A Small, Fast, Cheap and Lightweight Diffusion Synthesized Deepfake Detection},
  author={Yewon Lim and Changyeon Lee and Aerin Kim and Oren Etzioni},
  journal={ArXiv},
  year={2024},
  volume={abs/2406.00856},
  url={https://api.semanticscholar.org/CorpusID:270210472}
}

@article{Wen2016LearningSS,
  title={Learning Structured Sparsity in Deep Neural Networks},
  author={Wei Wen and Chunpeng Wu and Yandan Wang and Yiran Chen and Hai Helen Li},
  journal={Neural Information Processing Systems},
  year={2016},
  volume={abs/1608.03665},
  url={https://api.semanticscholar.org/CorpusID:2056019}
}

@article{Fang2023DepGraphTA,
  title={DepGraph: Towards Any Structural Pruning},
  author={Gongfan Fang and Xinyin Ma and Mingli Song and Michael Bi Mi and Xinchao Wang},
  journal={2023 IEEE/CVF Conference on Computer Vision and Pattern Recognition (CVPR)},
  year={2023},
  pages={16091-16101},
  url={https://api.semanticscholar.org/CorpusID:256390345}
}

@article{Wang2020NeuralPV,
  title={Neural Pruning via Growing Regularization},
  author={Huan Wang and Can Qin and Yulun Zhang and Yun Raymond Fu},
  journal={International Conference on Learning Representations},
  year={2020},
  volume={abs/2012.09243},
  url={https://api.semanticscholar.org/CorpusID:229297917}
}

@article{Liu2017LearningEC,
  title={Learning Efficient Convolutional Networks through Network Slimming},
  author={Zhuang Liu and Jianguo Li and Zhiqiang Shen and Gao Huang and Shoumeng Yan and Changshui Zhang},
  journal={2017 IEEE International Conference on Computer Vision (ICCV)},
  year={2017},
  pages={2755-2763},
  url={https://api.semanticscholar.org/CorpusID:5993328}
}

@inproceedings{Lee2020LayeradaptiveSF,
  title={Layer-adaptive Sparsity for the Magnitude-based Pruning},
  author={Jaeho Lee and Sejun Park and Sangwoo Mo and Sungsoo Ahn and Jinwoo Shin},
  booktitle={International Conference on Learning Representations},
  year={2020},
  url={https://api.semanticscholar.org/CorpusID:234358843}
}

@article{Rombach2021HighResolutionIS,
  title={High-Resolution Image Synthesis with Latent Diffusion Models},
  author={Robin Rombach and A. Blattmann and Dominik Lorenz and Patrick Esser and Bj{\"o}rn Ommer},
  journal={2022 IEEE/CVF Conference on Computer Vision and Pattern Recognition (CVPR)},
  year={2021},
  pages={10674-10685},
  url={https://api.semanticscholar.org/CorpusID:245335280}
}

@article{Zhang2023AddingCC,
  title={Adding Conditional Control to Text-to-Image Diffusion Models},
  author={Lvmin Zhang and Anyi Rao and Maneesh Agrawala},
  journal={2023 IEEE/CVF International Conference on Computer Vision (ICCV)},
  year={2023},
  pages={3813-3824},
  url={https://api.semanticscholar.org/CorpusID:256827727}
}

@article{Das2024BDfreshwaterfishAI,
  title={BD-freshwater-fish: An image dataset from Bangladesh for AI-powered automatic fish species classification and detection toward smart aquaculture},
  author={Pranajit Kumar Das and Md. Abu Kawsar and Puspendu Biswas Paul and Md. Abdullah Al Mamun Hridoy and Md. Sanowar Hossain and Sabyasachi Niloy},
  journal={Data in Brief},
  year={2024},
  volume={57},
  url={https://api.semanticscholar.org/CorpusID:274089964}
}

@article{Wei2024AGL,
  title={A Green Learning Approach to Spoofed Speech Detection},
  author={Chengwei Wei and Runqi Pang and C.-C. Jay Kuo},
  journal={ICASSP 2024 - 2024 IEEE International Conference on Acoustics, Speech and Signal Processing (ICASSP)},
  year={2024},
  pages={12956-12960},
  url={https://api.semanticscholar.org/CorpusID:268530843}
}

@article{Kuo2022GreenLI,
  title={Green Learning: Introduction, Examples and Outlook},
  author={C.-C. Jay Kuo and Azad M. Madni},
  journal={ArXiv},
  year={2022},
  volume={abs/2210.00965},
  url={https://api.semanticscholar.org/CorpusID:252683160}
}

@article{Zhu2022RGGIDAR,
  title={RGGID: A Robust and Green GAN-Fake Image Detector},
  author={Yao Zhu and Xinyu Wang and Ronald Salloum and Hong-Shuo Chen and C.-C. Jay Kuo},
  journal={APSIPA Transactions on Signal and Information Processing},
  year={2022},
  url={https://api.semanticscholar.org/CorpusID:255255923}
}

@article{Zhu2021APixelHopAG,
  title={A-PixelHop: A Green, Robust and Explainable Fake-Image Detector},
  author={Yao Zhu and Xinyu Wang and Hong-Shuo Chen and Ronald Salloum and C.-C. Jay Kuo},
  journal={ICASSP 2022 - 2022 IEEE International Conference on Acoustics, Speech and Signal Processing (ICASSP)},
  year={2021},
  pages={8947-8951},
  url={https://api.semanticscholar.org/CorpusID:243848088}
}

@misc{chen2024gift,
  title     = {{GIFT: A Green Image Forgery Tracker}},
  author    = {Chen, Hong-Shuo and Zhu, Yao and Yu, Chee-An and Salloum, Ronald and Kuo, C.-C. Jay},
  year      = {2024},
  note      = {12 pages, posted: 22 Mar 2024},
  doi       = {10.2139/ssrn.4768871}
}

\end{document}